\documentclass{article}

\usepackage[final]{nips_2018_arxiv}

\usepackage{graphicx}
\usepackage{subcaption}

\usepackage[utf8]{inputenc} 
\usepackage[T1]{fontenc}    
\usepackage{hyperref}       
\usepackage{url}            
\usepackage{booktabs}       
\usepackage{amsfonts}       
\usepackage{amsmath}
\usepackage{nicefrac}       
\usepackage{microtype}      

\usepackage{cancel}			 
\usepackage{mathtools}

\usepackage{xcolor}
\usepackage{xspace}
\definecolor{mygreen}{rgb}{0.2, 0.7, 0.2}
\definecolor{myorange}{rgb}{0.9, 0.5, 0.0}
\definecolor{myyellow}{rgb}{0.7, 0.7, 0.0}

\newcommand{\R}{\mathbb{R}}

\newcommand{\E}{\mathrm{E}}
\newcommand{\Var}{\mathrm{Var}}

\newcommand{\norm}{\mathcal{N}}
\newcommand{\dirichlet}{\mathrm{Dir}}
\newcommand{\categorical}{\mathrm{Cat}}
\newcommand{\gammadist}{\mathrm{Gamma}}
\newcommand{\lognorm}{\mathrm{Lognormal}}

\newcommand{\fvect}{\mathbf{f}}

\newcommand{\xvect}{\mathbf{x}}
\newcommand{\yvect}{\mathbf{y}}

\newcommand{\alphavect}{\boldsymbol{\alpha}}

\newcommand{\pivect}{\boldsymbol{\pi}}

\newcommand{\muvect}{\boldsymbol{\mu}}

\newcommand{\sigmavect}{\boldsymbol{\sigma}}

\newcommand{\name}[1]{{\textsc{#1}}\xspace}

\newcommand{\eeg}{\name{eeg}}
\newcommand{\magic}{\name{magic}}
\newcommand{\htru}{\name{htru2}}
\newcommand{\miniboo}{\name{miniboo}}
\newcommand{\coverbin}{\name{coverbin}}
\newcommand{\letter}{\name{letter}}
\newcommand{\drive}{\name{drive}}
\newcommand{\mocap}{\name{mocap}}

\newcommand{\susy}{\name{susy}}

\newcommand{\gp}{\name{gp}}
\newcommand{\gps}{\textsc{gp}s\xspace}

\newcommand{\gpr}{\textsc{gpr}\xspace}
\newcommand{\gprc}{\textsc{gpr (platt)}\xspace}
\newcommand{\gpc}{\textsc{gpc}\xspace}
\newcommand{\gpd}{\textsc{gpd}\xspace}

\newcommand{\krr}{\textsc{krr}\xspace}
\newcommand{\nkrr}{\textsc{nkrr}\xspace}

\newcommand{\svms}{\textsc{svm}s\xspace}

\newcommand{\rbf}{\name{rbf}}

\newcommand{\ece}{\name{ece}}
\newcommand{\mnll}{\name{mnll}}

\usepackage{lipsum}

\title{Dirichlet-based Gaussian Processes\\ for Large-scale Calibrated Classification}

\author{
  Dimitrios~Milios\\
  EURECOM\\
  Sophia Antipolis, France \\
  \texttt{milios@eurecom.fr} \\
  \And
  Raffaello~Camoriano \\
  University of Genoa \\
  Genoa, Italy \\
  \texttt{raffaello.camoriano@iit.it} \\
  \AND
  Pietro~Michiardi \\
  EURECOM \\
  Sophia Antipolis, France \\
  \texttt{Pietro.Michiardi@eurecom.fr} \\
  \And
  Lorenzo~Rosasco \\
  University of Genoa \\ LCSL, IIT \& MIT \\
  \texttt{lrosasco@mit.edu} \\
  \And
  Maurizio~Filippone \\
  EURECOM \\
  Sophia Antipolis, France \\
  \texttt{maurizio.filippone@eurecom.fr} \\
}

\begin{document}
\maketitle

\begin{abstract}

In this paper, we study the problem of deriving fast and accurate classification algorithms with uncertainty quantification. 
Gaussian process classification provides a principled approach, but the corresponding computational burden is hardly sustainable in large-scale problems and devising efficient alternatives is a challenge. In this work, we investigate if and how 
Gaussian process regression directly applied to the classification labels  can be used to tackle this question.  
While in this case training time is remarkably  faster, predictions need be calibrated for classification and uncertainty estimation. To this aim, we propose a novel approach 
based on  interpreting  the labels as the output of a Dirichlet distribution.  
Extensive experimental results show  that the proposed approach provides  essentially the same accuracy and uncertainty quantification of  Gaussian process classification  while requiring only a fraction of computational resources.

\end{abstract}

\section{Introduction}
\label{sec:intro}

Classification is a classic machine learning task.
While the most basic performance measure is classification accuracy, in practice assigning a calibrated confidence to the  predictions is often crucial \cite{Flach16}.
For example, in image classification providing classification predictions with a calibrated score  is important to avoid making over-confident decisions \cite{Guo17,Kendall17,Kurakin17}. 
Several classification algorithms that output a continuous score are not necessarily calibrated (e.g., support vector machines \cite{Platt99b}). 
Popular ways to calibrate classifiers use a validation set to learn a transformation of their output score that recovers calibration; these include Platt scaling \cite{Platt99b} and isotonic regression \cite{Zadrozny02}. Calibration can also be   achieved  if  a sensible loss function is employed \cite{Kull17}, for example  the logistic/cross-entropy loss,  and it is known to be positively impacted  if the classifier is well regularized \cite{Guo17}.  

Bayesian approaches provide a natural framework to tackle these kinds of questions, since quantification of uncertainty is of primary interest. 
In particular,  Gaussian Processes Classification (\gpc) \cite{Rasmussen06,Williams98,Hensman15b} combines the flexibility of Gaussian Processes (\gps) \cite{Rasmussen06} and the regularization stemming from their probabilistic nature, with the use of the correct likelihood for classification, that is Bernoulli or multinomial for binary or multi-class classication, respectively.  While we are not aware of empirical studies on the calibration properties of \gpc,  our results confirm the intuition that \gpc is actually calibrated.  The most severe drawback of \gpc, however, is that it requires carrying out several matrix factorizations, making it unattractive for large-scale problems. 

In this paper, we study the question of whether Gaussian process approaches can be made efficient  to find accurate and well-calibrated classification rules. 
A simple idea is to use Gaussian process regression directly on classification labels. 
This idea is quite common in  non-probabilistic approaches \cite{Suykens99,Rifkin03} and can be grounded from a decision theoretic point of view.
Indeed,  the Bayes' rule minimizing the expected least squares is the the expected conditional probability,  which in classification is directly related to the conditional probabilities of each class, see e.g. \cite{Shawe-Taylor04,bartlett2006}. 
Directly regressing the labels leads  to fast training and excellent classification accuracies \cite{Rudi17,Lu14,Huang14}.
However, the corresponding predictions are not calibrated for uncertainty quantification. The question is then if calibration can be achieved while retaining speed. 

The main contribution of our work is the proposal of a transformation of the classification labels, which turns the original problem into a regression problem without compromising calibration.   For \gps, this has the enormous advantage of bypassing the need for expensive posterior approximations, leading to a method that is as fast as a simple regression of the original labels.
The proposed method is based on the interpretation of the labels as the output of a Dirichlet distribution, so we name it Dirichlet-based \gp classification (\gpd).
Through an extensive experimental validation, including large-scale classification tasks, we demonstrate that \gpd is calibrated and competitive in performance with state-of-the-art \gpc.

\section{Related work}
\label{sec:related:work}

\paragraph{Calibration of classifiers:}
Platt scaling \cite{Platt99b} is a popular method to calibrate the output score of classifiers, as well as isotonic regression \cite{Zadrozny02}.
More recently, Beta calibration \cite{Kull17} and temperature scaling \cite{Guo17} have been proposed to extend the class of possible transformations and reduce the parameterization of the transformation, respectively.
It is established that binary classifiers are calibrated when they employ the logistic loss;
this is a direct consequence of the fact that the appropriate model for Bernoulli distributed variables is the one associated with this loss \cite{Kull17}. 
The extension to multi-class problems yields the so-called cross-entropy loss, which corresponds to the multinomial likelihood.
Not necessarily, however, the right loss makes classifiers well calibrated; recent works on calibration of convolutional neural networks for image classification show that depth negatively impacts calibration due to the introduction of a large number of parameters to optimize, and that regularization is important to recover calibration \cite{Guo17}.

\paragraph{Kernel-based classification:} 
Performing regression on classification labels is also known as least-squares classification \cite{Suykens99,Rifkin03}.
We are not aware of works that study \gp-based least-squares classification in depth; we could only find a few comments on it in \cite{Rasmussen06} (Sec.~6.5). 
\gpc is usually approached assuming a latent process, which is given a \gp prior, that is transformed into a probability of class labels through a suitable squashing function \cite{Rasmussen06}.
Due to the non-conjugacy between the \gp prior and the non-Gaussian likelihood, applying standard Bayesian inference techniques in \gpc leads to analytical intractabilities, and it is necessary to resort to approximations.
Standard ways to approximate computations include the Laplace Approximation \cite{Williams98} and Expectation Propagation \cite{Minka01}; see, e.g., \cite{Kuss05,Nickisch08} for a detailed review of these methods. 
More recently, there have been advancements in works that extend ``sparse'' \gp approximations \cite{Titsias09} to classification \cite{Hensman15} in order to deal with the issues of scalability with the number of observations.
All these approaches require iterative refinements of the posterior distribution over the latent \gp, and this implies expensive algebraic operations at each iteration.

\section{Background}

Consider a multi-class classification problem. 
Given a set of $N$ training inputs $X = \{\xvect_1,\dots,\xvect_N\}$ and their corresponding labels $Y = \{\yvect_1,\dots,\yvect_N\}$, with one-hot encoded classes denoted by the vectors $\yvect_i$, a classifier produces a predicted label $\fvect(\xvect_*)$ as function of any new input $\xvect_*$.

In the literature, calibration is assessed through the \emph{Expected Calibration Error} (\ece) \cite{Guo17}, which is the average of the absolute difference between accuracy and confidence:
\begin{equation}
\ece = \sum_{m=1}^{M} \frac{|X_m|}{|X_*|} \left\vert\text{acc}(\fvect(X_m), Y_m) - \text{conf}(\fvect, X_m)\right\vert,
\end{equation}
where the test set $X_*$ is divided into disjoint subsets $\{X_{1}, \dots, X_{M}\}$, each corresponding to a given level of confidence $\text{conf}(\fvect, X_m)$ predicted by the classifier $\fvect$, while $\text{acc}(\fvect(X_m), Y_m)$ is the classification accuracy of $\fvect$ measured on the $m$-th subset.
Other metrics used in this work to characterize the quality of a classifier are the \emph{mean negative log-likelihood} (\mnll) of the test set under the model, and the error rate on the test set.
All metrics are defined so that lower values are better.

\subsection{Kernel methods for classification}

\paragraph{\gp classification (\gpc)} \gp-based classification is defined by the following abstract steps:
\begin{enumerate}
\item A Gaussian prior is placed over a \emph{latent} function $f(\xvect)$.
The \gp prior is characterized by the mean function $\mu(\xvect)$ and the covariance function $k(\xvect, \xvect')$.
The observable (non-Gaussian) prior is obtained by transforming through a sigmoid function so that the sampled functions produce proper probability values.
In the multi-class case, we consider $C$ independent priors over the vector of functions $\fvect = [f_1,\dots,f_C]^\top$; transformation to proper probabilities is achieved by applying the softmax function $\sigmavect(\fvect)$ \footnote{Softmax function $\sigmavect(\fvect)$ s.t. $\sigma(\fvect)_j = \exp(f_j) / \sum_{c = 1}^C \exp(f_c)$ for $j = 1,...C$}.
\item The observed labels $\yvect$ are associated with a categorical likelihood with probability components $p(y_c \mid \fvect) = \sigmavect(\fvect(\xvect))_c$, for any $c \in \{1,\dots,C\}$.
\item The latent posterior is obtained by means of Bayes' theorem.
\item The latent posterior is transformed via $\sigmavect(\fvect)$, to obtain a distribution over class probabilities.
\end{enumerate}

Throughout this work, we consider $\mu(\xvect) = 0$ and $k(\xvect, \xvect') = a^2 \exp\left(-\frac{(\xvect - \xvect')^2}{2 l^2}\right)$. 
This choice of covariance function is also known as the \rbf kernel; it is characterized by the $a^2$ and $l$ hyper-parameters, interpreted as the \gp marginal variance and length-scale, respectively.
The hyper-parameters are commonly selected my maximizing the marginal likelihood of the model.

The major computational challenge of \gpc can be identified in Step 3 described above.
The categorical likelihood implies that the posterior stochastic process is not Gaussian and it cannot be calculated analytically.
Therefore, different approaches resort to Gaussian approximations of the likelihood so that the resulting approximate Gaussian posterior has the following form:
\begin{equation}
p(\fvect \mid X, Y) \approx q(\fvect \mid X, Y) \sim p(\fvect) \norm(\tilde{\muvect}, \tilde{\Sigma}).
\end{equation}
For example, in Expectation Propagation (EP, \cite{Minka01}) $\tilde{\muvect}$ and $\tilde{\Sigma}$ are determined by the \emph{site parameters}, which are learned via an iterative process.
In the variational approach of \cite{Hensman15}, $\tilde{\muvect}$ and $\tilde{\Sigma}$ are the variational parameters of the approximate Gaussian likelihoods.
Despite being successful, approaches like these contribute significantly to the computational cost of classification, as they introduce a large number of parameters that need to be optimized.
In this work, we explore a more straightforward Gaussian approximation to the likelihood that requires no significant computational overhead.

\paragraph{\gp regression (\gpr)  on classification labels}
A simple way to bypass the problem induced by categorical likelihoods is to perform least-squares regression on the labels by ignoring their discrete nature.
This implies considering a Gaussian likelihood $p(\yvect \mid \fvect) = \norm(\fvect, \sigma_n^2)$, where $\sigma_n^2$ is the observation noise variance.
It is well-known that if the observed labels are $0$ and $1$, then the function $\fvect$ that minimizes the mean squared error converges to the true class probabilities in the limit of infinite data \cite{Rasmussen2006}. 
Nevertheless, by not squashing $\fvect$ through a softmax function, we can no longer guarantee that the resulting distribution of functions will lie within 0 and 1.
For this reason, additional calibration steps are required (i.e.\ Platt scaling).

\paragraph{Kernel Ridge Regression (\krr) for classification}
The idea of directly regressing labels is quite common when \gp estimators are applied within a frequentist context \cite{Rifkin03}.
Here they are typically derived from a non-probabilistic perspective based on empirical risk minimization, and the corresponding approach is dubbed Kernel Ridge Regression \cite{hastie2001elements}.
Taking this perspective, two comments can be made.
The first is  that the noise and covariance parameters are viewed as regularization parameters that need to be tuned,  typically by cross-validation.
In our experiments, we compare this method with a canonical \gpr approach.
The second comment  is that regressing labels with least squares can be justified from a decision theoretic point of view.
The Bayes' rule minimizing the expected least squares is the regression function  (the expected conditional probability), which in binary classification is proportional to the conditional probability of one of the two classes \cite{bartlett2006} (similar reasoning applies to multi-class classification \cite{mroueh2012multiclass, baldassarre2012multi}).
From this perspective, one could expect a least squares estimator to be self-calibrated,
however this is typically not the case in practice, a feature imputed to  the limited number of points and the choice of function models. 
In the following we breifly present Platt scaling, a simple and effective post-hoc calibration method which can be seamlessly applied to both \gpr- and \krr-based learning pipelines to obtain a calibrated model.

\paragraph{Platt scaling} Platt scaling \cite{Platt99b} is an effective approach to perform post-hoc calibration for different types of classifiers, such as \svms \cite{NiculescuMizil2005} and neural networks \cite{Guo17}.
Given a decision function $f$, which is the result of a trained binary classifier, the class probabilities are given by the sigmoid transformation $\pi(\xvect) = \sigma(a f(\xvect) + b)$, where $a$ and $b$ are optimised over a separate validation set, so that the resulting model best explains the data.
Although this parametric form may seem restrictive, Platt scaling has been shown to be effective for a wide range of classifiers \cite{NiculescuMizil2005}.

\subsection{A note on calibration properties}


We advocate that two components are critical for well-calibrated classifiers: \emph{regularization} and the \emph{cross-entropy loss}.
Previous work indicates that regularization has a positive effect on calibration \cite{Guo17}.
Also, classifiers that rely on the cross-entropy loss are reported to be well-calibrated \cite{NiculescuMizil2005}.
This form of loss function is equivalent to the negative Bernoulli log-likelihood (or categorical in the multi-class case), which is the proper interpretation of classification outcomes.

\begin{figure}
\centering
\includegraphics[width=\linewidth]{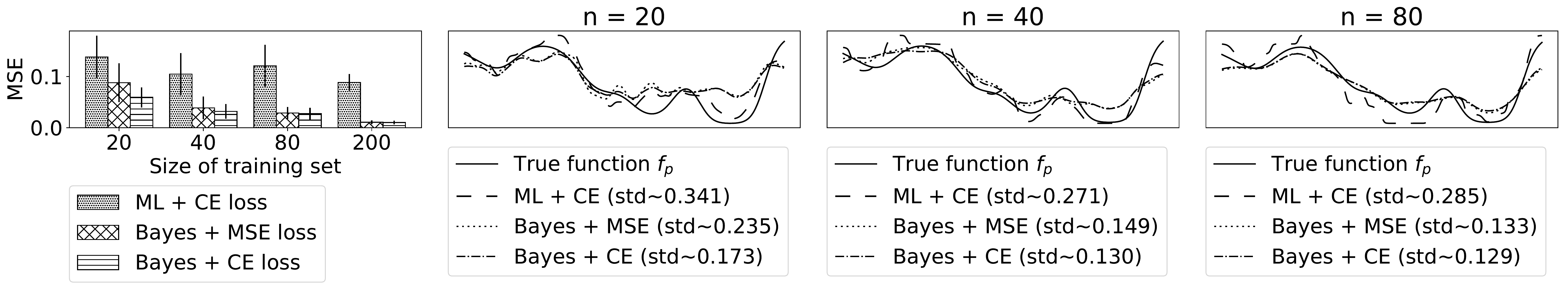}
\caption{Convergence of classifiers with different loss functions and regularization properties. Left: summary of the mean squared error (MSE) from the true function $f_p$ for 1000 randomly sampled training sets of different size; the Bayesian CE-based classifier is characterized by smaller variance even when the number of training inputs is small. Right: demonstration of how the averaged classifiers approximate the true function for different training sizes.}
\label{fig:loss_convergence}
\end{figure}

In Figure \ref{fig:loss_convergence}, we demonstrate the effects of regularization and cross-entropy empirically: we summarize classification results on four synthetic datasets of increasing size.
We assume that each class label is sampled from a Bernoulli distribution with probability given by the unknown function $f_p: \R^d \rightarrow [0, 1]$ with input space of dimensionality $d$.
For a classifier to be well-calibrated, it should accurately approximate $f_p$.
We fit three kinds of classifiers: a maximum likelihood (ML) classifier that relies on cross entropy loss (CE), a Bayesian classifier with MSE loss (i.e. \gpr classification), and finally a Bayesian classifier that relies on CE (i.e.\ \gpc).
We report the averages over 1000 iterations and the average standard deviation.
The Bayesian classifiers that rely on the cross entropy loss converge to the true solution at a faster rate, and they are characterized by smaller variance.

Although performing \gpr on the labels induces regularization through the prior, the likelihood model is not appropriate.
One possible solution is to employ meticulous likelihood approximations such as EP or variational \gp classification \cite{Hensman15}, alas at an often prohibitive computational cost, especially for considerably large datasets.
In the section that follows, we introduce a methodology that combines the best of both worlds.
We propose to perform \gp regression on labels transformed in such a way that a less crude approximation of the categorical likelihood is achieved.

\section{\gp regression on transformed Dirichlet variables}
\label{sec:gpd}

There is an obvious defect in \gp-based least-squares classification: each point is associated with a Gaussian likelihood, which is not the appropriate noise model for Bernoulli-distributed variables.
Instead of approximating the true non-Gaussian likelihood, we propose to transform the labels in a latent space where a Gaussian approximation to the likelihood is more sensible.


For a given input, the goal of a Bayesian classifier is to estimate the distribution over its class probability vector; such a distribution is naturally represented by a Dirichlet-distributed random variable.
More formally, in a $C$-class classification problem each observation $\yvect$ is a sample from a categorical distribution $\categorical(\pivect)$.
The objective is to infer the class probabilities $\pivect = [\pi_1,\dots, \pi_C]^\top$, for which we use a Dirichlet model: $\pivect \sim \dirichlet(\alphavect)$. 
In order to fully describe the distribution of class probabilities, we have to estimate the concentration parameters $\alphavect = [\alpha_1,\dots, \alpha_C]^\top$.
Given an observation $\yvect$ such that $y_k = 1$, our best guess for the values of $\alphavect$ will be: $\alpha_k=1+\alpha_\epsilon$ and $\alpha_i=\alpha_\epsilon, \forall i\neq k$.
Note that it is necessary to add a small quantity $0 < \alpha_\epsilon \ll 1$, so as to have valid parameters for the Dirichlet distribution.
Intuitively, we implicitly induce a Dirichlet prior so that before observing a data point we have the probability mass shared equally across $C$ classes; we know that we should observe exactly one count for a particular class, but we do not know which one.
Most of the mass is concentrated on the corresponding class when $\yvect$ is observed.
This practice can be thought of as the categorical/Bernoulli analogue of the noisy observations in \gp regression.
The likelihood model is:
\begin{equation}
p(\yvect \mid \alphavect) = \categorical(\pivect)
, \quad \mathrm{where\ } \pivect \sim \dirichlet(\alphavect) \text{.}
\end{equation}
It is well-known that a Dirichlet sample can be generated by sampling from $C$ independent Gamma-distributed random variables with shape parameters $\alpha_i$ and rate $\lambda=1$; realizations of the class probabilities can be generated as follows:
\begin{equation}
\pi_i = \frac{x_i}{\sum_{c=1}^{C} x_c}
, \quad \mathrm{where\ } x_i \sim \gammadist(\alpha_i, 1)
\label{eq:dirichlet_samples}
\end{equation}
Therefore, the noisy Dirichlet likelihood assumed for each observation translates to $C$ independent Gamma likelihoods with shape parameters either $\alpha_i = 1+\alpha_\epsilon$, if $y_i=1$, or $\alpha_i = \alpha_\epsilon$ otherwise.

In order to construct a Gaussian likelihood in the log-space, we approximate each Gamma-distributed $x_i$ with $\tilde{x}_i \sim \lognorm(\tilde{y}_i, \tilde{\sigma}_i^2)$, which has the same mean and variance (i.e. moment matching):
\begin{align*}
\E[x_i] = \E[\tilde{x}_i]      \Leftrightarrow \alpha_i &= \exp(\tilde{y}_i + \tilde{\sigma}_i^2 / 2)\\
\Var[x_i] = \Var[\tilde{x}_i]  \Leftrightarrow \alpha_i &= \left(\exp(\tilde{\sigma}_i^2) - 1\right) \exp(2\tilde{y}_i + \tilde{\sigma}_i^2)
\end{align*}
Thus, for the parameters of the normally distributed logarithm we have:
\begin{align}
\tilde{y}_i        = \log \alpha_i - \tilde{\sigma}_i^2 / 2, \quad\quad\quad\quad
\tilde{\sigma}_i^2 = \log(1/\alpha_i + 1)
\end{align}


Note that this is the first approximation to the likelihood that we have employed so far.
One could argue that a log-normal approximation to a Gamma-distributed variable is reasonable, although it is not perfect for small values of the shape parameter $\alpha_i$.
However, the most important implication is that we can now consider a Gaussian likelihood in the log-space.
Assuming a vector of latent processes $\fvect = [f_1,\dots,f_C]^\top$, we have:
\begin{equation}
p(\tilde{y}_i \mid \fvect) = \norm(f_i, \tilde{\sigma}_i^2),
\end{equation}
where the class label is now denoted by $\tilde{y}_i$ in the transformed logarithmic space.
It is important to note that the noise parameter $\tilde{\sigma}_i^2$ is different for each observation; we have a {\em heteroskedastic} regression model.
In fact, the $\tilde{\sigma}_i^2$ values (as well as $\tilde{y}_i$) solely depend on the Dirichlet pseudo-count assumed in the prior, which only has two possible values.
Given this likelihood approximation, it is straightforward to place a \gp prior over $\fvect$ and evaluate the posterior over $C$ latent processes exactly.

\textbf{Remark:} In the binary classification case, we still have to perform regression on two latent processes.
The use of heteroskedastic noise model implies that one latent process is not a mirrored version of the other (see Figure~\ref{fig:dirichlet_example}), contrary to \gpc.

\subsection{From \gp posterior to Dirichlet variables}

The obtained \gp posterior emulates the logarithm of a stochastic process with Gamma marginals that gives rise to the Dirichlet posterior distributions.
It is straightforward to sample from the posterior log-normal marginals, which should behave as Gamma-distributed samples to generate posterior Dirichlet samples as in Equation \eqref{eq:dirichlet_samples}.
It is easy to see that this corresponds to a simple application of the softmax function on the posterior \gp samples.
The expectation of class probabilities will be:
\begin{equation}
\E[\pi_{i,*} \mid X] = \int \frac{\exp(f_{i,*})}{\sum_j \exp(f_{j,*})} \, p(f_{i,*} \mid X) \, d\fvect_*
\label{eq:softmax_expectation}
\end{equation}
which can be approximated by sampling from the Gaussian posterior $p(f_{i,*} \mid X)$.

\begin{figure}
\includegraphics[width=\linewidth]{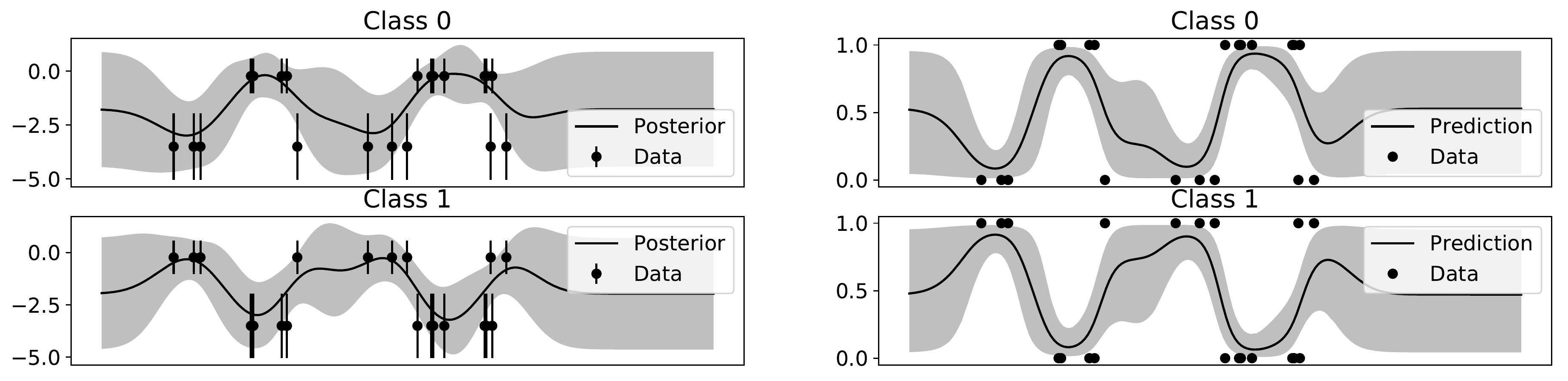}
\caption{Example of Dirichlet regression for a one-dimensional binary classification problem. Left: the latent \gp posterior for class ``0'' (top) and class ``1'' (bottom). Right: the transformed posterior through softmax for class ``0'' (top) and class ``1'' (bottom).}
\label{fig:dirichlet_example}
\end{figure}

Figure \ref{fig:dirichlet_example} is an example of Dirichlet regression for a one-dimensional binary classification problem.
The left-side panels demonstrate how the \gp posterior approximates the transformed data; the error bars represent the standard deviation for each data-point.
Notice, that the posterior for class ``0'' (top) is not a mirror image of class ``1'' (bottom), because of the different noise terms assumed for each latent process.
The right-side panels show results in the original output space, after applying softmax transformation; as expected in the binary case, one posterior process is a mirror image of the other.

\subsection{Optimizing the Dirichlet prior $\alpha_\epsilon$}

The performance of Dirichlet-based classification is affected by the choice of $\alpha_\epsilon$, in addition to the usual \gp hyperparameters.
As $\alpha_\epsilon$ approaches zero, $\alpha_i$ converges to either $1$ or $0$.
It is easy to see that for the transformed ``1'' labels we have $\tilde{\sigma}^2_i = \log 2$ and $\tilde{y}_i = \log(1/\sqrt{2})$ in the limit.
The transformed ``0'' labels, however, converge to infinity, and so do their variances.
The role of $\alpha_\epsilon$ is to make the transformed labels finite, so that it is possible to perform regression.
The smaller the $\alpha_\epsilon$ is, the further the transformed labels will be apart, but at the same time, the variance for the ``0'' label will be larger.

By increasing $\alpha_\epsilon$, the transformed labels of different classes tend to be closer.
The marginal log-likelihood tends to be larger, as it is easier for a zero-mean \gp prior to fit the data.
However, this behavior is not desirable for classification purposes. 
For this reason, the Gaussian marginal log-likelihood in the transformed space is not appropriate to determine the optimal value for $\alpha_\epsilon$.

\begin{figure}
\centering
	\begin{subfigure}{0.24\textwidth}
	\includegraphics[width=\linewidth]{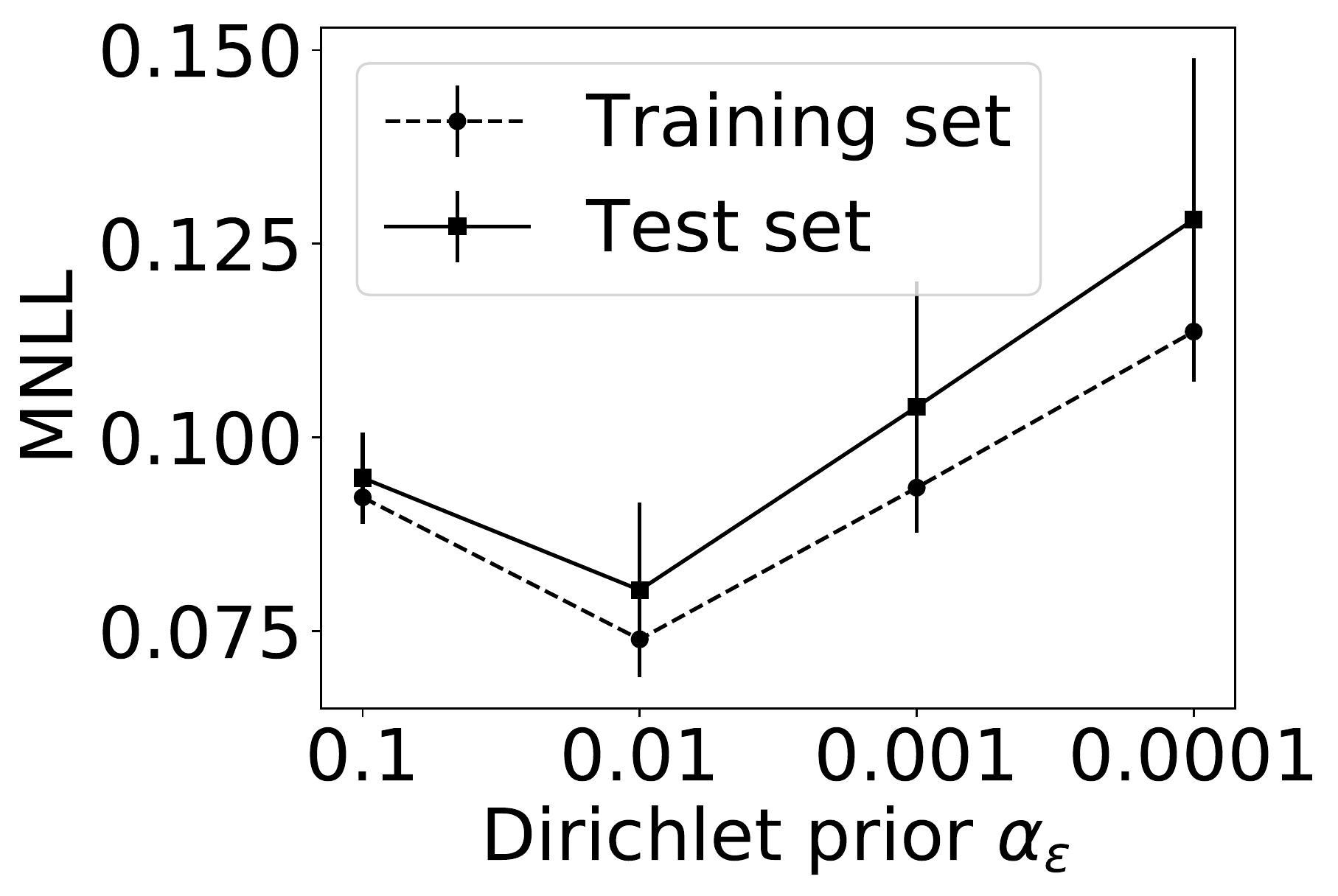}
	\caption{\htru}
	\end{subfigure}
	\begin{subfigure}{0.24\textwidth}
	\includegraphics[width=\linewidth]{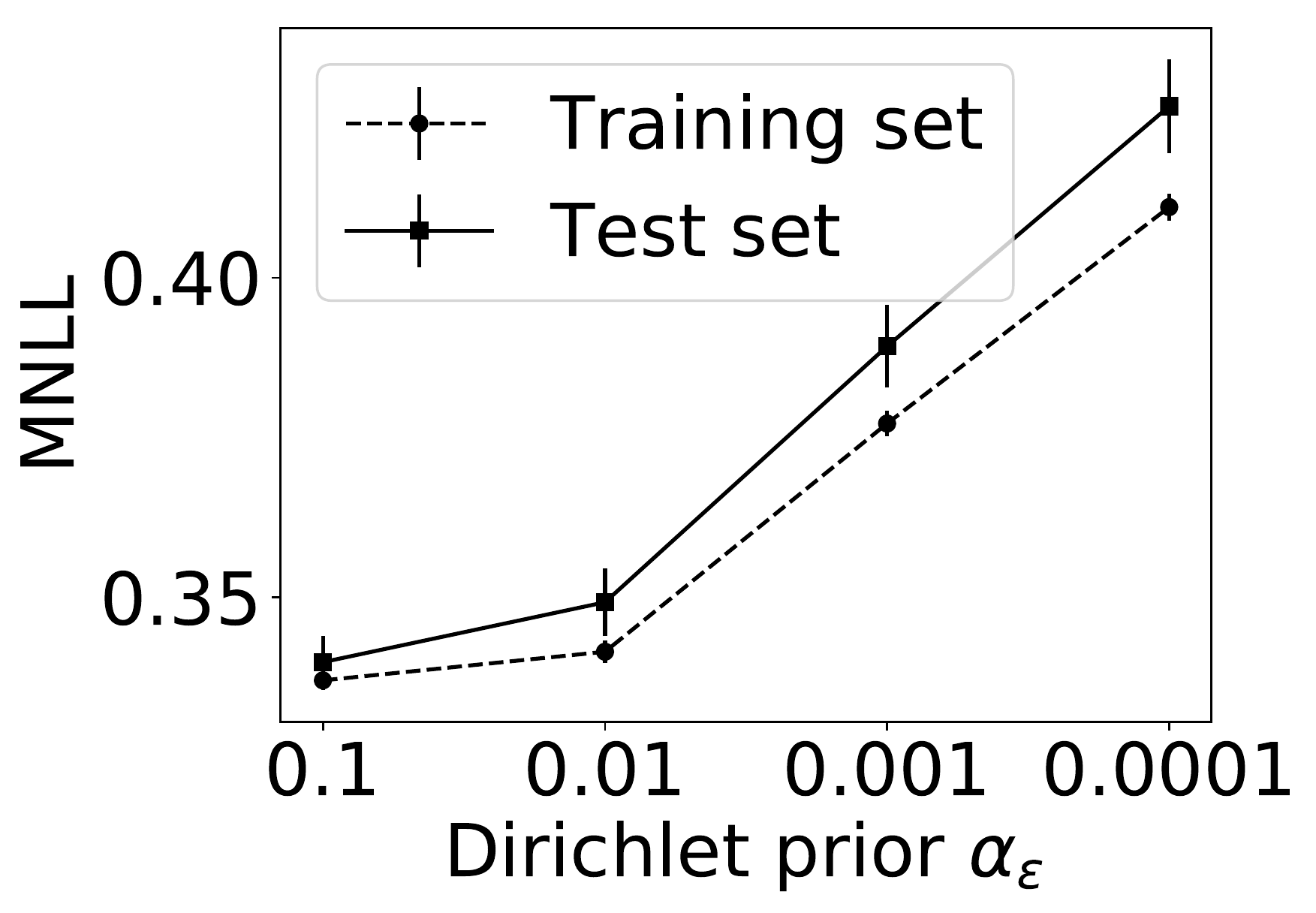}
	\caption{\magic}
	\end{subfigure}
	\begin{subfigure}{0.24\textwidth}
	\includegraphics[width=\linewidth]{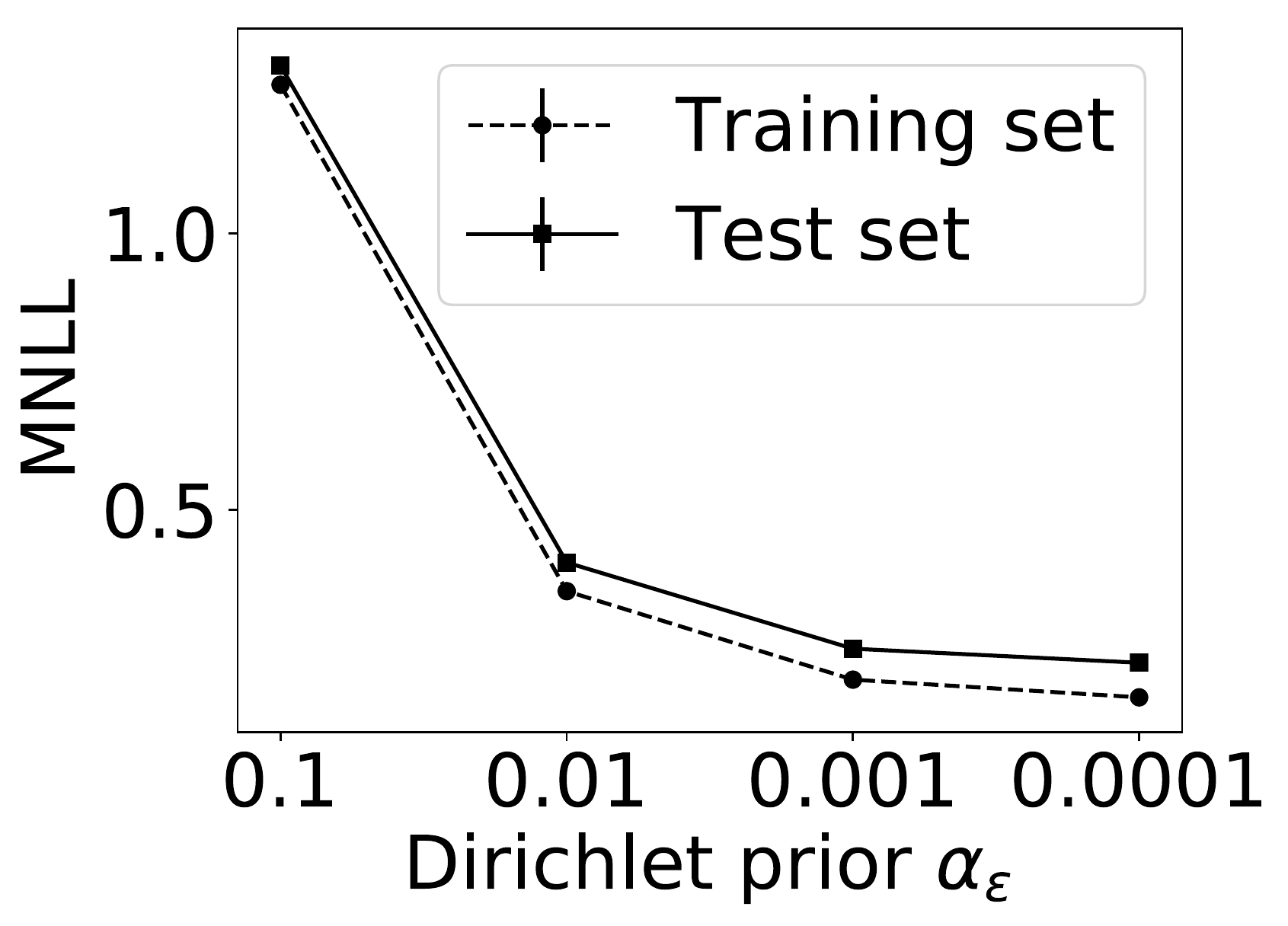}
	\caption{\letter}
	\end{subfigure}
	\begin{subfigure}{0.24\textwidth}
	\includegraphics[width=\linewidth]{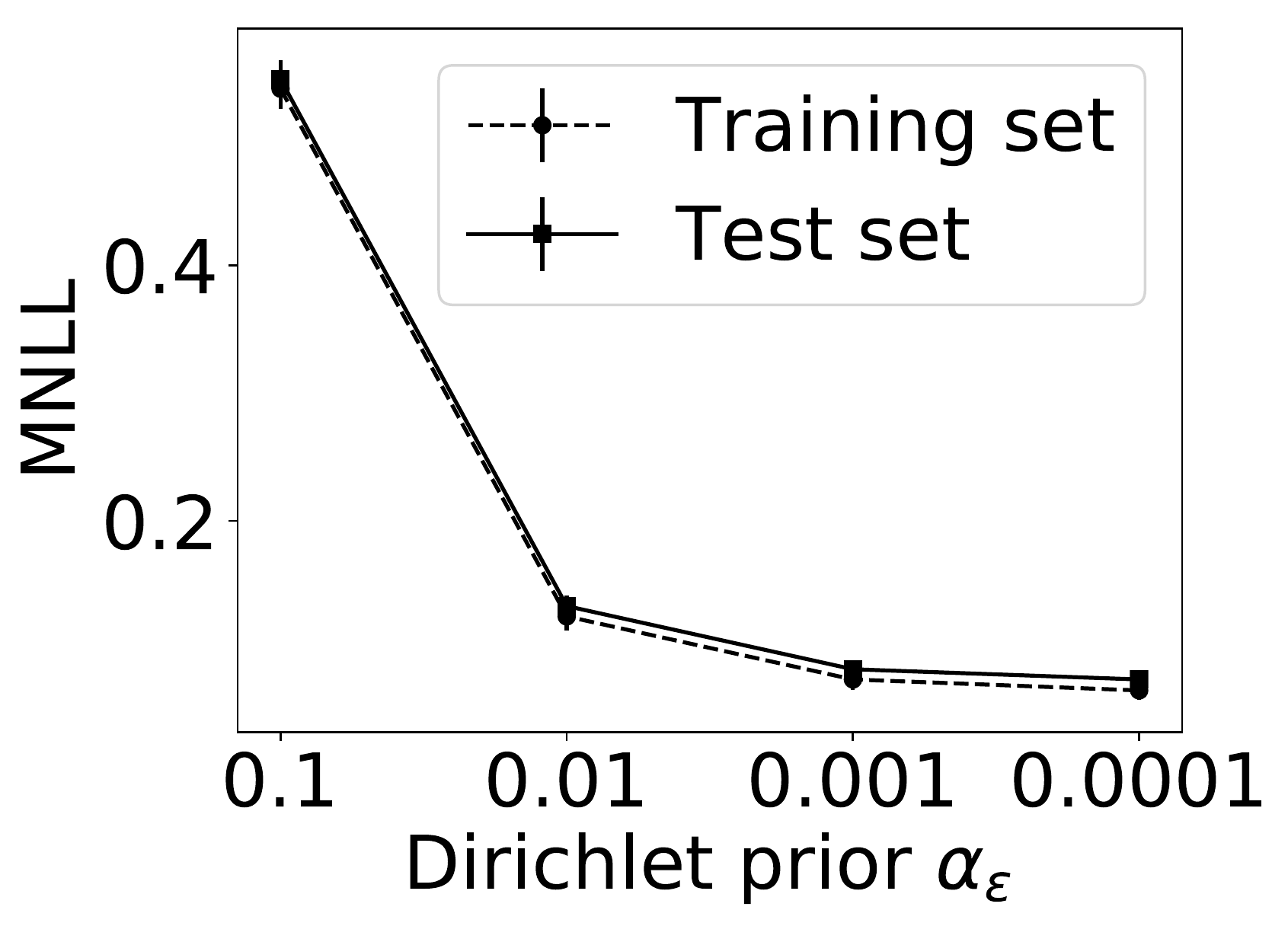}
	\caption{\drive}
	\end{subfigure}
\caption{Exploration of $\alpha_\epsilon$ for 4 different datasets with respect to the \mnll metric.}
\label{fig:explore_dprior}
\end{figure}

Figure \ref{fig:explore_dprior} demonstrates the effect of $\alpha_\epsilon$ on classification accuracy, as reflected by the \mnll metric.
Each subfigure corresponds to a different dataset; \mnll is reported for different choices of $\alpha_\epsilon$ between $0.1$ and $0.001$.
As a general remark, it appears that there is no globally optimal $\alpha_\epsilon$ parameter across datasets.
However, the reported training and testing \mnll curves appear to be in agreement regarding the optimal choice for $\alpha_\epsilon$.
We therefore propose to select the $\alpha_\epsilon$ value that minimizes the \mnll on the training data.

\section{Experiments}
\label{sec:experiments}

We experimentally evaluate the methodologies discussed on the datasets outlined in Table \ref{tab:datasets}.
For the implementation of \gp-based models, we use and extend the algorithms available in the GPFlow library \cite{GPflow17}.
More specifically, for \gpc we make use of variational sparse \gp \cite{Hensman15b}, while for \gpr we employ sparse variational \gp regression \cite{Titsias09}.
The latter is also the basis for our \gpd implementation: we apply adjustments so that heteroskedastic noise is admitted, as dictated by the Dirichlet mapping.
Concerning \krr, in order to scale it up to large-scale problems we use a subsampling-based variant named Nystr\"om \krr (\nkrr) \cite{conf/icml/SmolaS00,Williams00b}. 
Nystr\"om-based approaches have been shown to achieve state-of-the-art accuracy on large-scale learning problems \cite{conf/nips/KumarMT09, conf/icml/SiHD14, Rudi15, camoriano2016nytro, rudi2017falkon}.
The number of inducing (subsampled) points used for each dataset is reported in Table \ref{tab:datasets}.

\begin{table}
\centering
\caption{Datasets used for evaluation, available from the UCI repository \cite{Asuncion07}.}
\label{tab:datasets}
\begin{tabular}{lrrrrr}
\toprule
Dataset 	& Classes 	& Training instances 	& Test instances 	& Dimensionality	& Inducing points \\
\midrule
\eeg		& 2		 	& 10980					& 4000		 		& 14				& 200 \\
\htru		& 2		 	& 12898					& 5000		 		& 8					& 200 \\
\magic		& 2		 	& 14020					& 5000		 		& 10				& 200 \\
\miniboo	& 2		 	& 120064				& 10000		 		& 50				& 400 \\
\coverbin	& 2		 	& 522910				& 58102		 		& 54				& 500 \\
\susy		& 2	 		& 4000000				& 1000000	 		& 18				& 200 \\
\letter		& 26	 	& 15000					& 5000		 		& 16				& 200 \\
\drive		& 11	 	& 48509					& 10000		 		& 48				& 500 \\
\mocap		& 5		 	& 68095					& 10000		 		& 37				& 500 \\
\bottomrule
\end{tabular}
\end{table}

The experiments have been repeated for 10 random training/test splits.
For each iteration, inducing points are chosen by applying k-means clustering on the training inputs.
Exceptions are \coverbin and \susy, for which we used 5 splits and inducing points chosen uniformly at random.
For \gpr we further split each training dataset: 80\% of which is used to train the model and the remaining 20\% is used for calibration with Platt scaling.
\nkrr uses an 80-20\% split for $k$-fold cross-validation and Platt scaling calibration, respectively.
For each of the datasets, the $\alpha_\epsilon$ parameter of \gpd was selected according to the training \mnll: we have $0.1$ for \coverbin, $0.001$ for \letter, \drive and \mocap, and $0.01$ for the remaining datasets.

In all experiments, we consider an isotropic \rbf kernel; the kernel hyperparameters are selected by maximizing the marginal likelihood for the \gp-based approaches, and by $k$-fold cross validation for \nkrr (with $k = 10$ for all datasets except from \susy, for which $k = 5$).
In the case of \gpr, we also optimize the noise variance jointly with all kernel parameters.

\begin{figure}
\centering
\includegraphics[width=\linewidth]{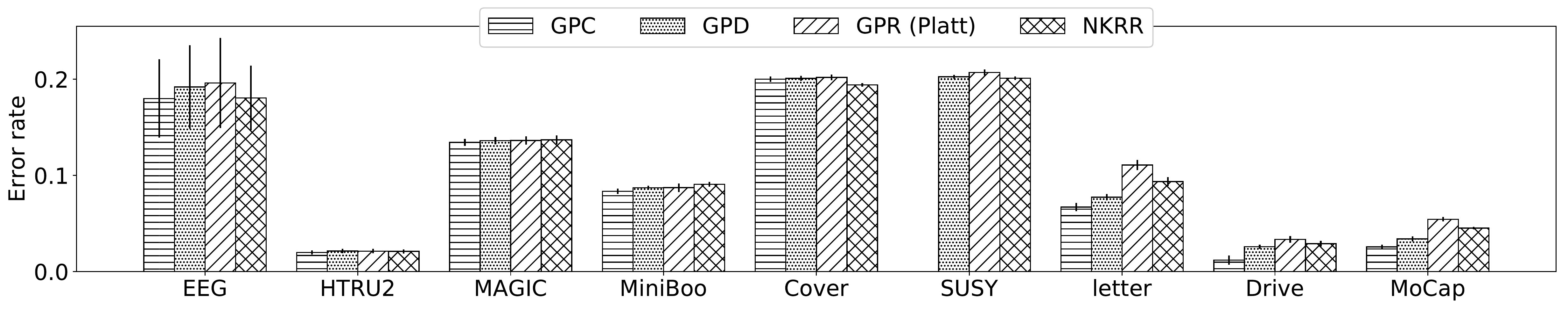}
\includegraphics[width=\linewidth]{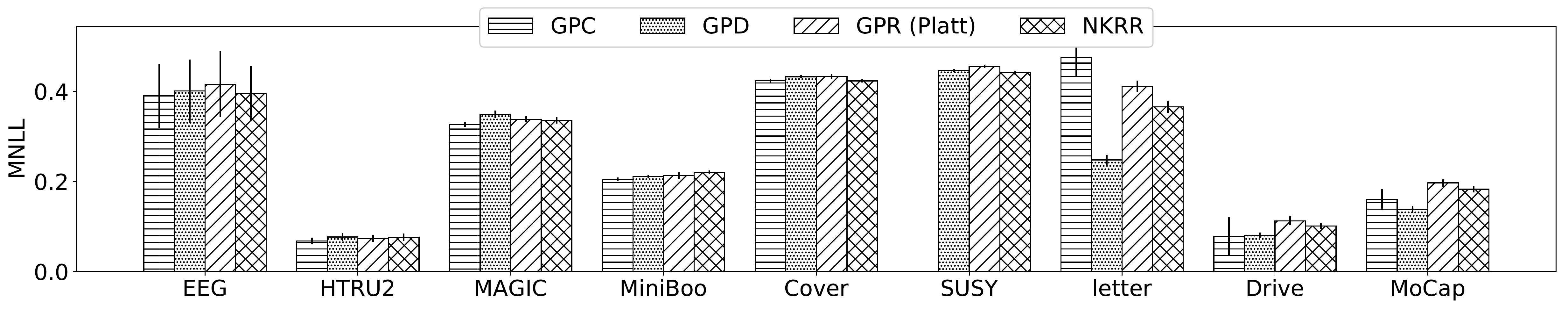}
\includegraphics[width=\linewidth]{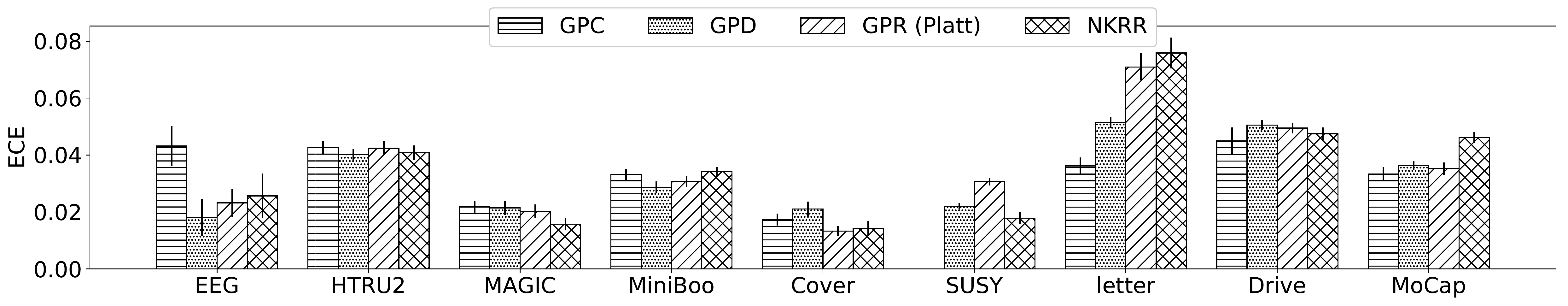}
\caption{Error rate, \mnll and \ece for the datasets considered in this work.}
\label{fig:metrics}
\end{figure}
\begin{figure}
\begin{subfigure}{0.32\textwidth}
\includegraphics[width=\linewidth]{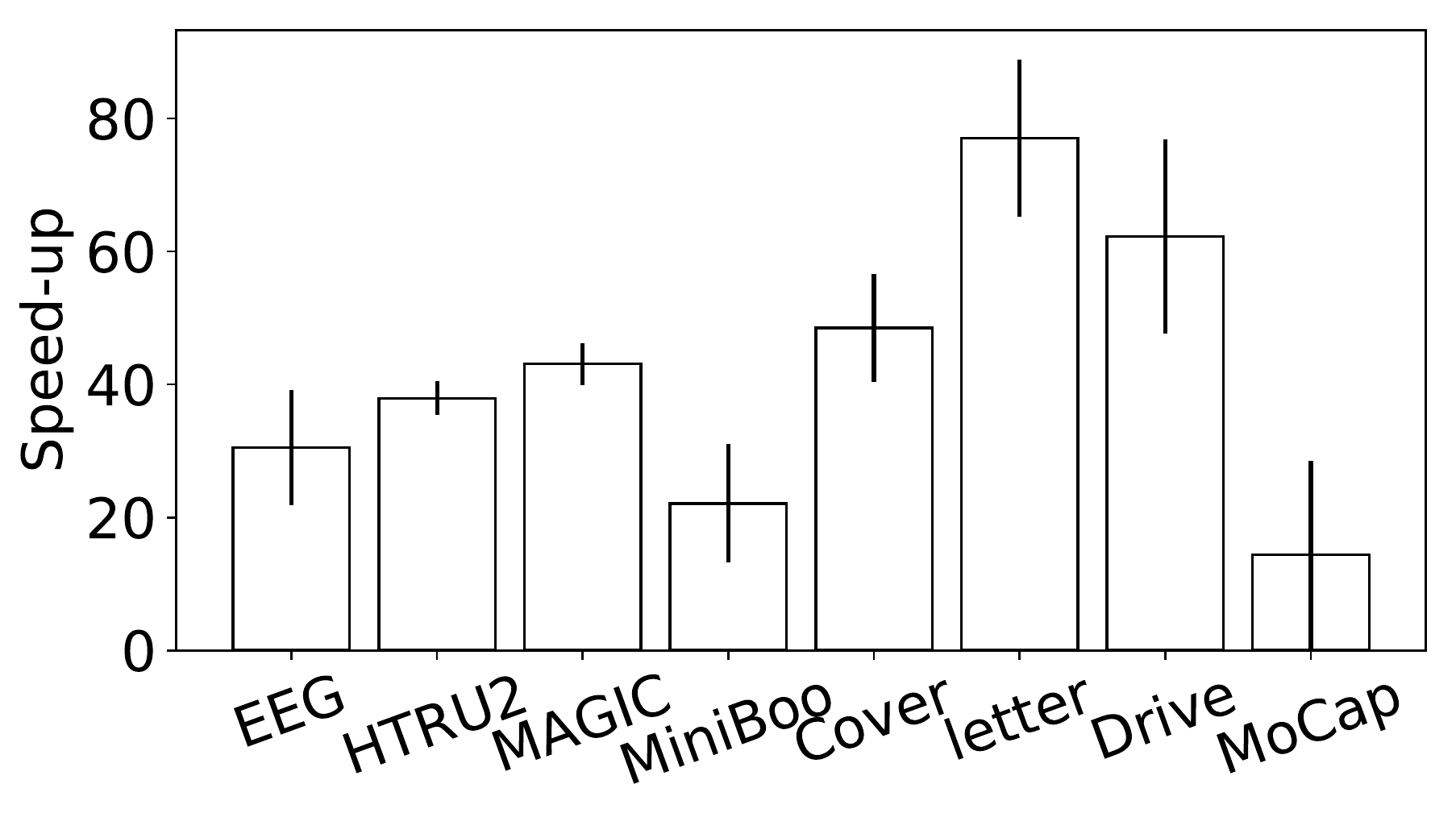}
\end{subfigure}
\begin{subfigure}{0.22\textwidth}
\includegraphics[width=\linewidth]{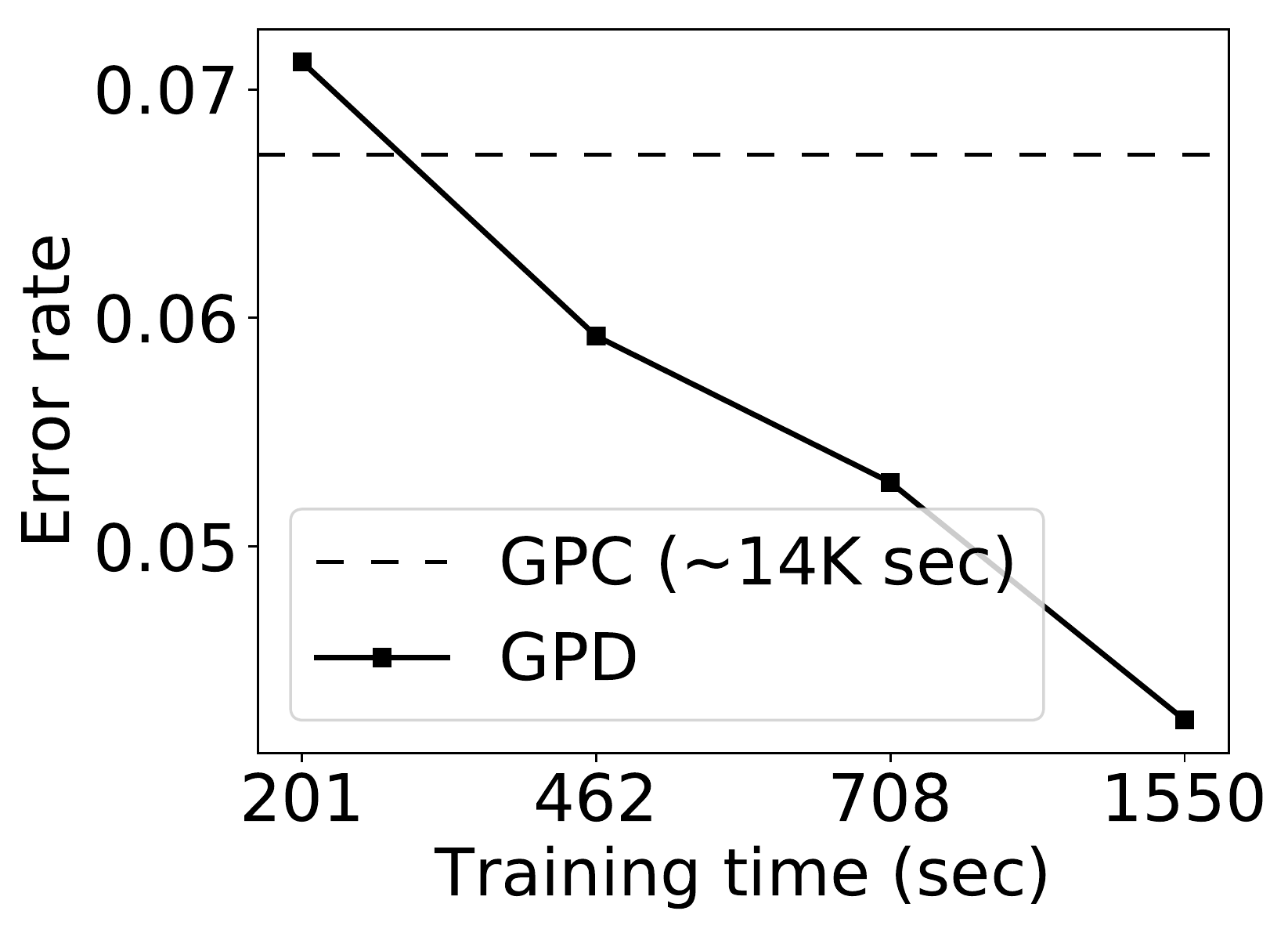}
\caption{\letter}
\end{subfigure}
\begin{subfigure}{0.22\textwidth}
\includegraphics[width=\linewidth]{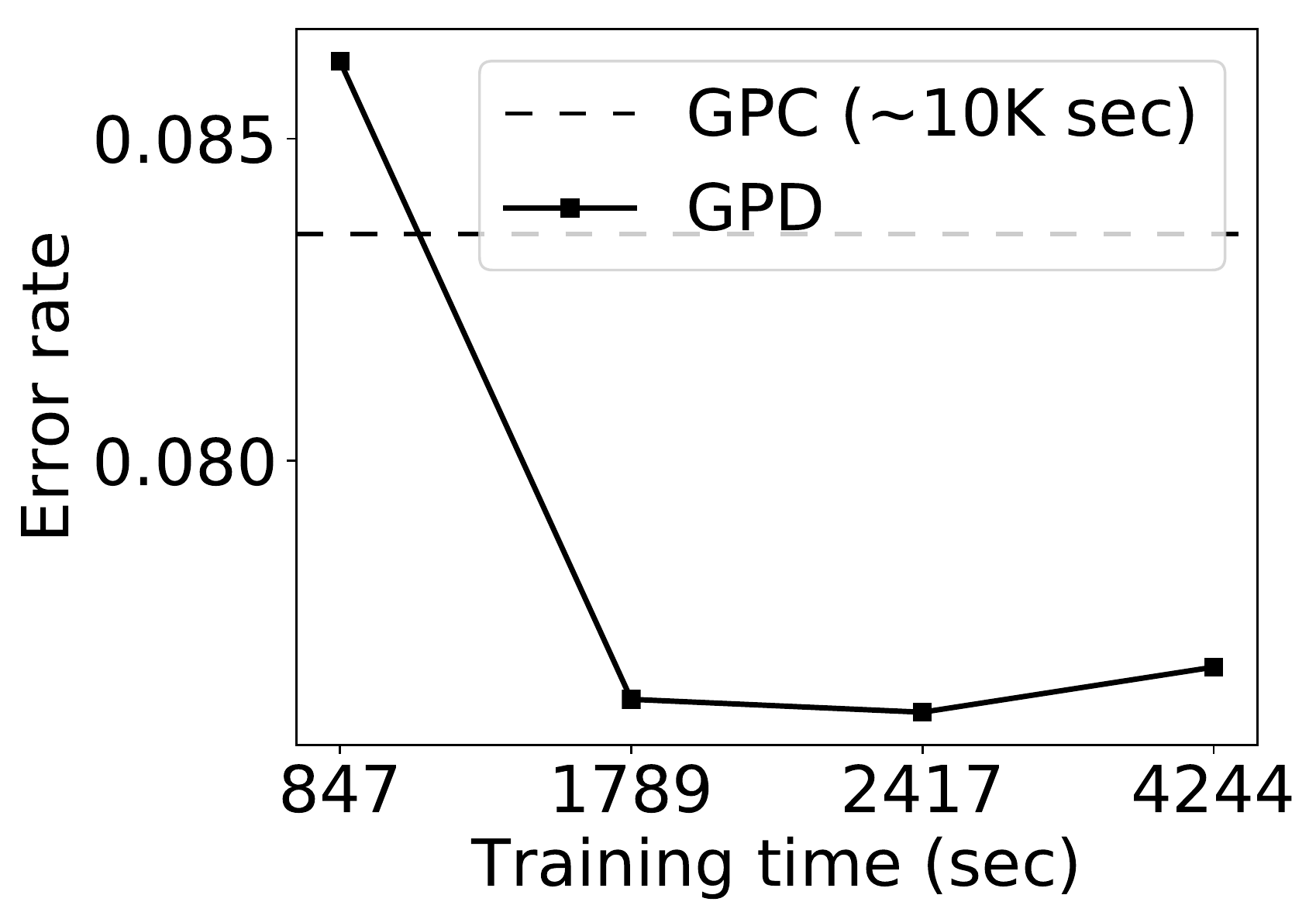}
\caption{\miniboo}
\end{subfigure}
\begin{subfigure}{0.22\textwidth}
\includegraphics[width=\linewidth]{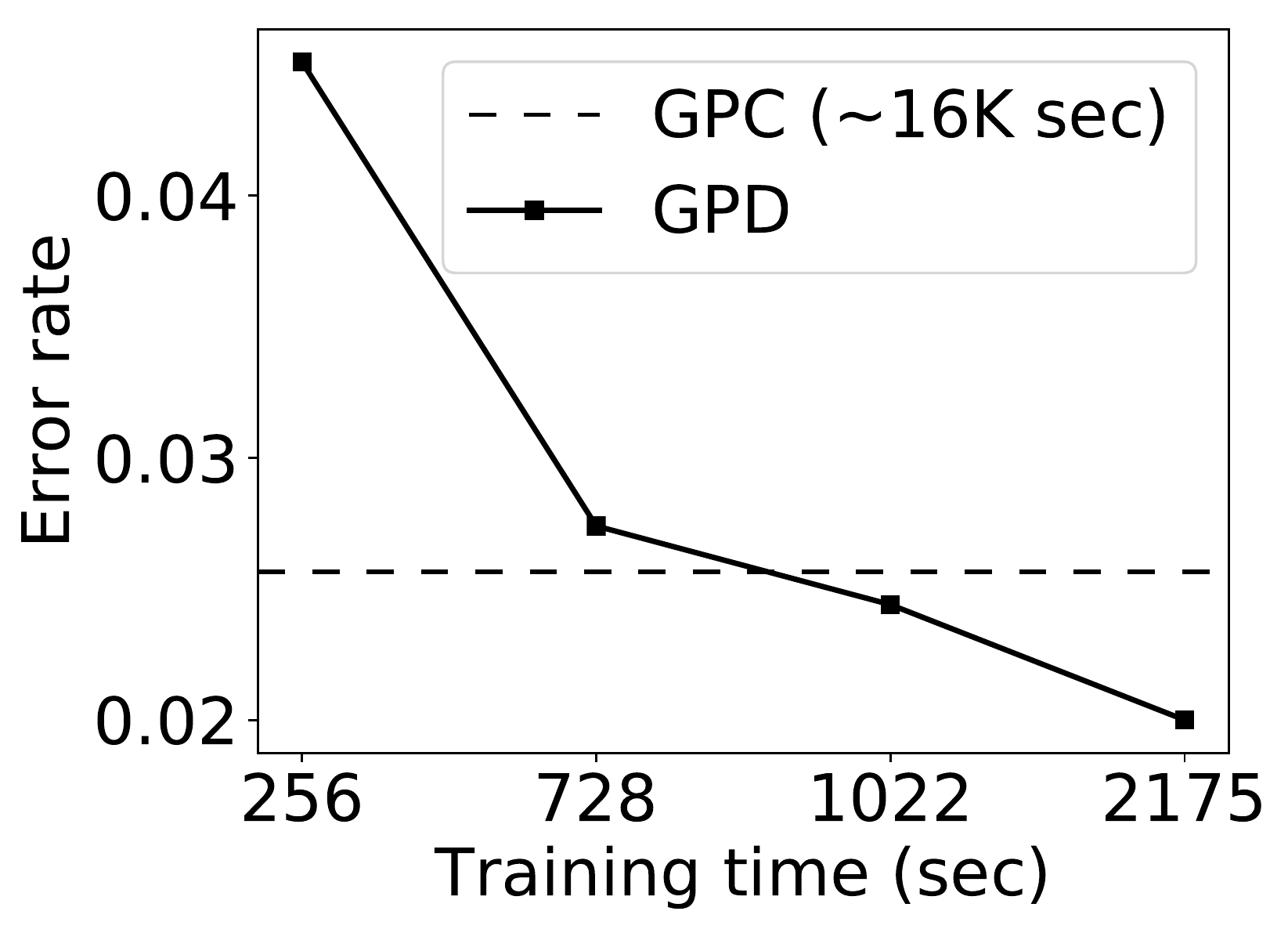}
\caption{\mocap}
\end{subfigure}
\caption{Left: Speed-up obtained by using \gpd as opposed to \gpc. Right: Error vs training time for \gpd as the number of inducing points is increased for three datasets. The dashed line represents the error obtained by \gpc using the same number of inducing points as the fastest \gpd listed.}
\label{fig:speedup}
\end{figure}

The performance of \gpd, \gpc, \gpr and \nkrr is compared in terms of various error metrics, including error rate, \mnll and \ece for a collection of datasets.
The error rate, \mnll and \ece values obtained are summarised in Figure~\ref{fig:metrics}. 
The \gpc method tends to outperform \gpr in most cases.
Regarding the \gpd approach, its performance tends to lie between \gpc and \gpr; in some instances classification performance is better than \gpc and \nkrr.
Most importantly, this performance is obtained at a fraction of the computational time required by the \gpc method.
Figure \ref{fig:speedup} summarizes the speed-up achieved by the used of \gpd as opposed to the variational \gp classification approach.

This dramatic difference in computational efficiency has some interesting implications regarding the applicability of \gp-based classification methods on large datasets.
\gp-based machine learning approaches are known to be computationally expensive; their practical application on large datasets demands the use of scalable methods to perform approximate inference.
The approximation quality of sparse approaches depends on the number (and the selection) of inducing points.
In the case of classification, the speed-up obtained by \gpd implies that the computational budget saved can be spent on a more fine-grained sparse \gp approximation.
In Figure \ref{fig:speedup}, we explore the effect of increasing the number of inducing points for three datasets: \letter, \miniboo and \mocap; we see that the error rate drops below the target \gpc with a fixed number of inducing points, and still at a fraction of the computational effort.

\section{Conclusions}
\label{sec:conslusions}

Most \gp-based approaches to classification in the literature are characterized by a meticulous approximation of the likelihood.
In this work, we experimentally show that such \gp classifiers tend to be well-calibrated, meaning that they estimate correctly classification uncertainty, as this is expressed through class probabilities.
Despite this desirable property, their applicability is limited to small/moderate size of datasets,
due to the high computational complexity of approximating the true posterior distribution.

Least-squares classification, which may be implemented either as \gpr or \krr, is an established practice for more scalable classification.
However, the crude approximation of a non-Gaussian likelihood with a Gaussian one has a negative impact on classification quality, especially as this is reflected by the calibration properties of the classifier.

Considering the strengths and practical limitations of \gps, we proposed a classification approach that is essentially an heteroskedastic \gp regression on a latent space induced by a transformation of the labels, which are viewed as Dirichlet-distributed random variables.
This allowed us to convert $C$-class classification to a problem of regression for $C$ latent processes with Gamma likelihoods.
We then proposed to approximate the Gamma-distributed variables with log-normal ones, and thus we achieved a sensible Gaussian approximation in the logarithmic space. 
Crucially, this can be seen as a pre-processing step, that does not have to be learned, unlike \gpc, where an accurate transformation is sought iteratively.
Our experimental analysis shows that Dirichlet-based \gp classification produces well-calibrated classifiers without the need for post-hoc calibration steps.
The performance of our approach in terms of classification accuracy tends to lie between properly-approximated \gpc and least-squares classification, but most importantly it is orders of magnitude faster than \gpc.

As a final remark, we note that the predictive distribution of the \gpd approach is different from that obtained by \gpc, as can be seen in the extended results in the appendix.
An extended characterization of the predictive distribution for \gpd is subject of future work.

\subsubsection*{Acknowledgments}
DM and PM are partially supported by KPMG.
RC would like to thank Luigi Carratino for the useful exchanges concerning large-scale experiments.
LR is funded by the Air Force project FA9550-17-1-0390 (European Office of Aerospace Research and Development) and the RISE project NoMADS - DLV-777826.
MF gratefully acknowledges support from the AXA Research Fund.

\small

\bibliographystyle{abbrv}

\bibliography{bibliography,filippone}

\appendix

\section{Extended calibration results}

\emph{Reliability diagrams} offer a visual representation of calibration properties, where accuracy is plotted as a function of confidence for the subsets $\{X_{1}, \dots, X_{M}\}$.
For a perfectly calibrated classifier, the accuracy function should be equal to the identity line, implying that $\text{accu}(X_m) = \text{conf}(X_m)$.
Large deviations from the identity line mean that the class probabilities are either underestimated or overestimated.

\begin{figure}[htp]
\centering
\includegraphics[width=0.24\linewidth]{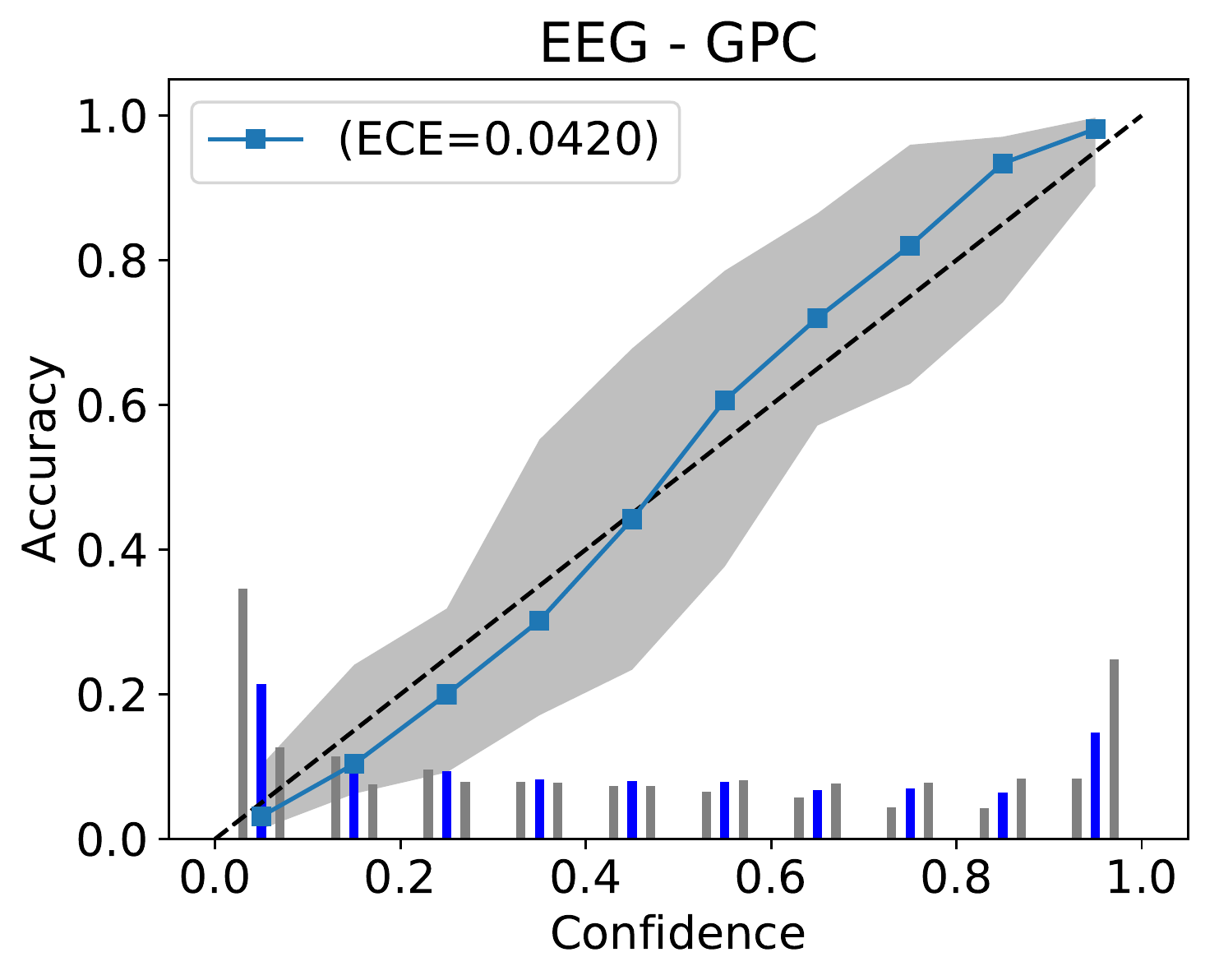}
\includegraphics[width=0.24\linewidth]{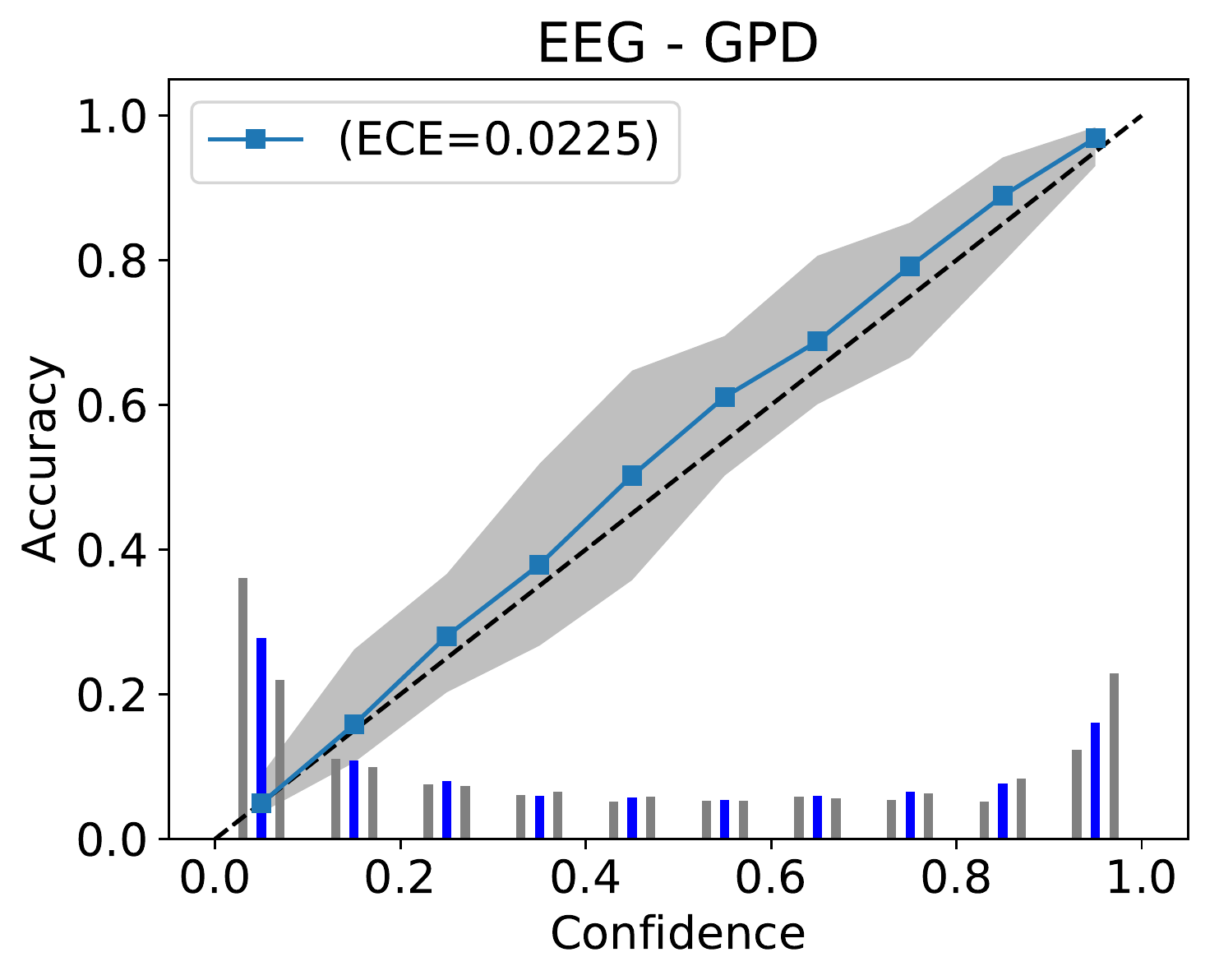}
\includegraphics[width=0.24\linewidth]{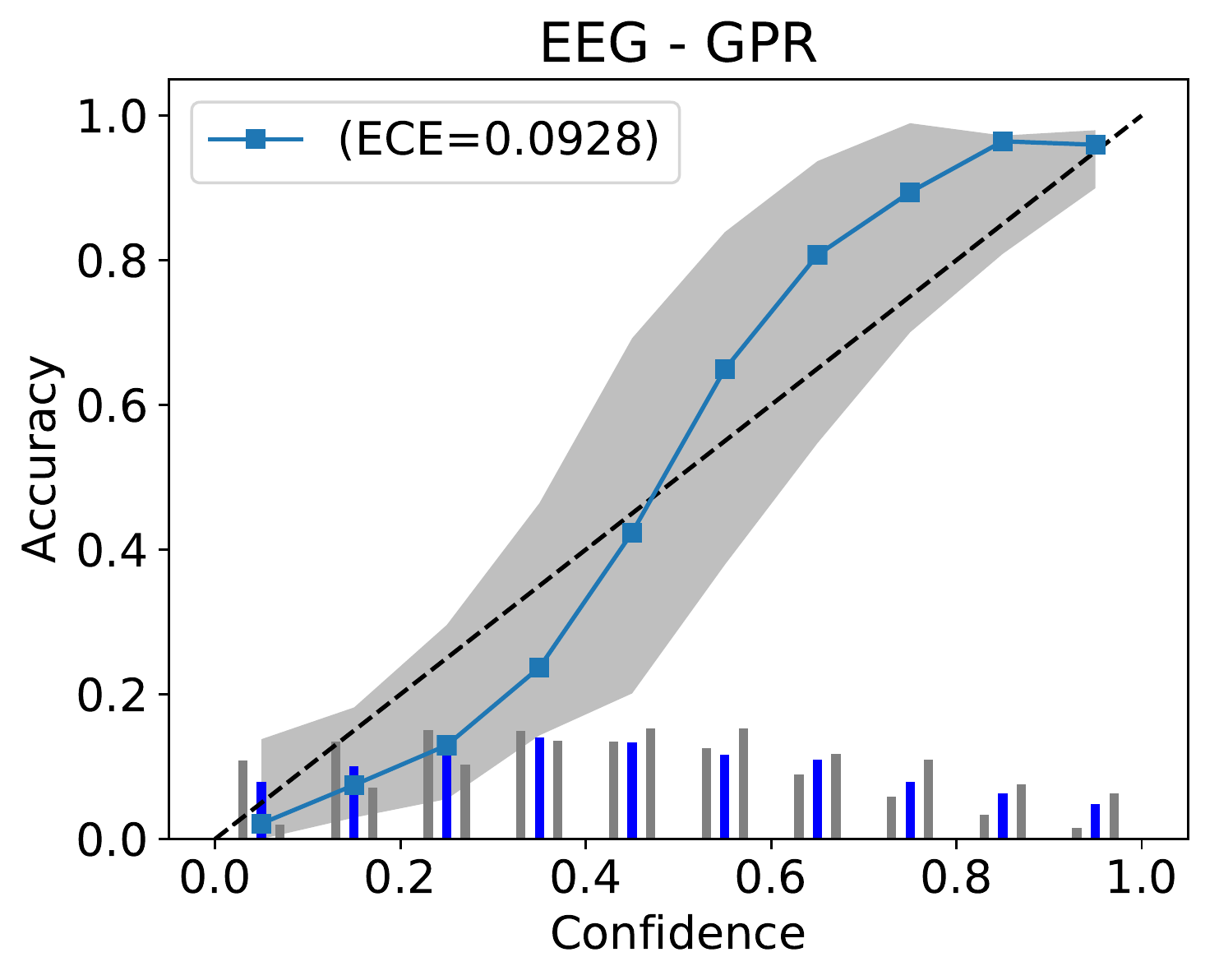}
\includegraphics[width=0.24\linewidth]{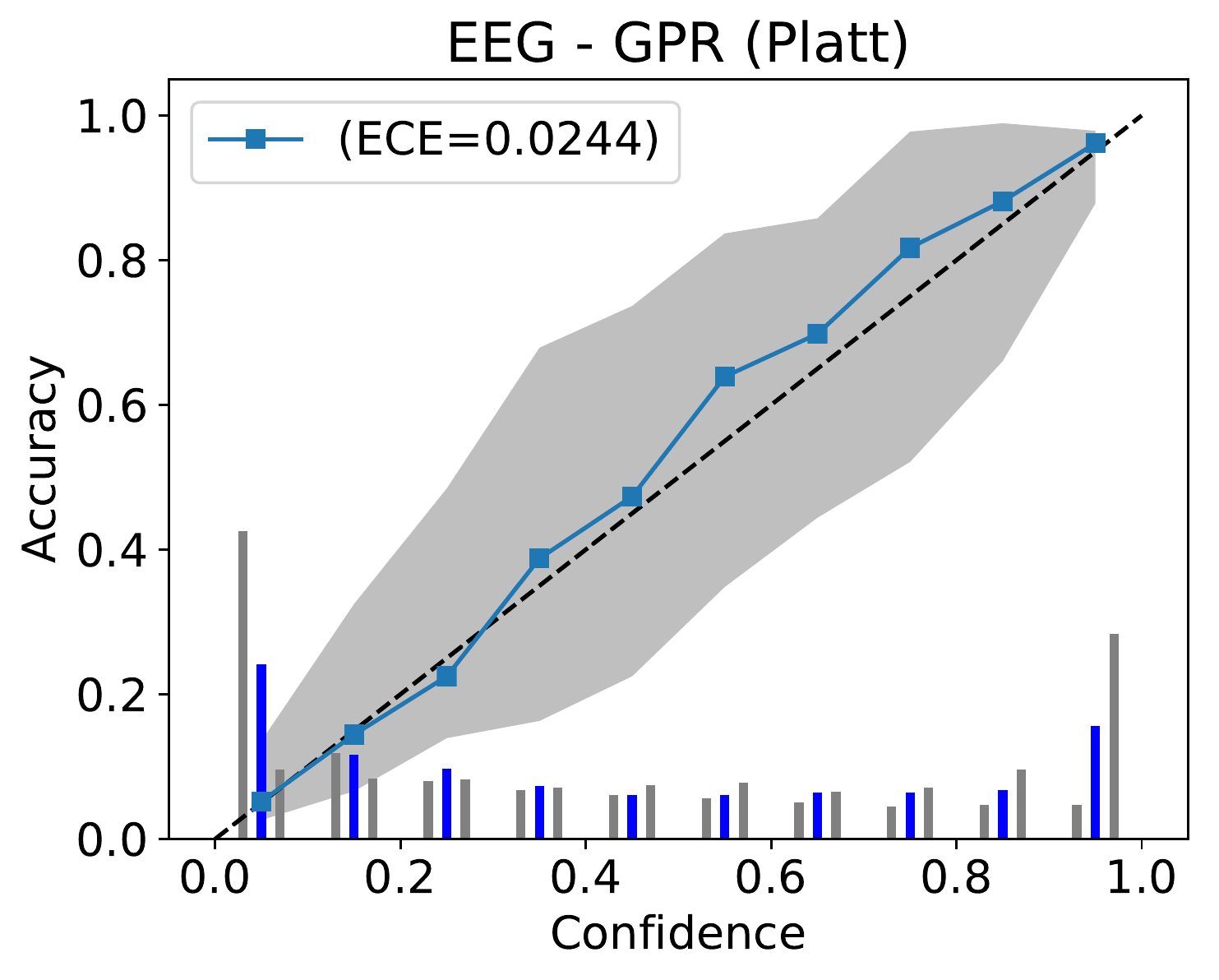}
\includegraphics[width=0.24\linewidth]{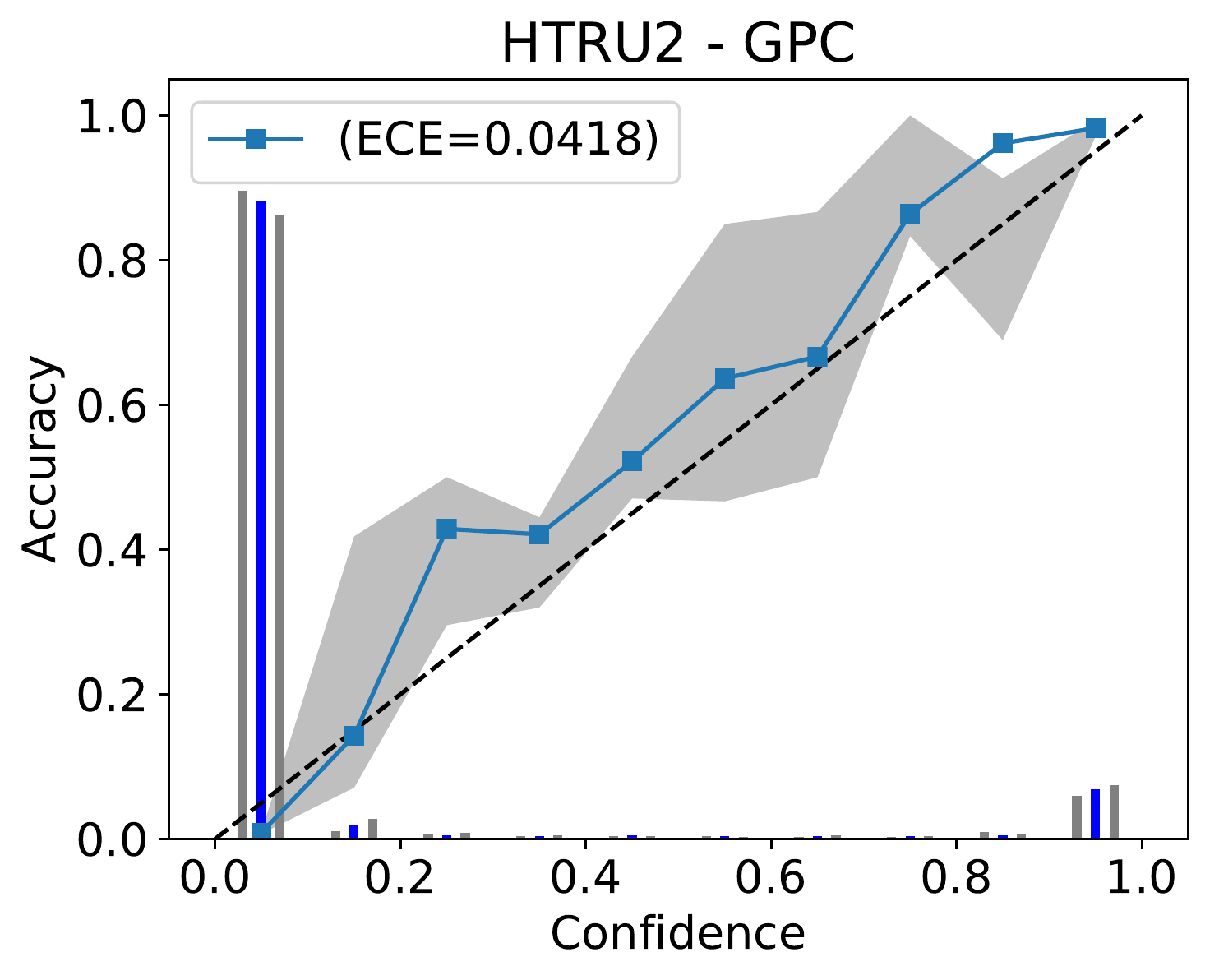}
\includegraphics[width=0.24\linewidth]{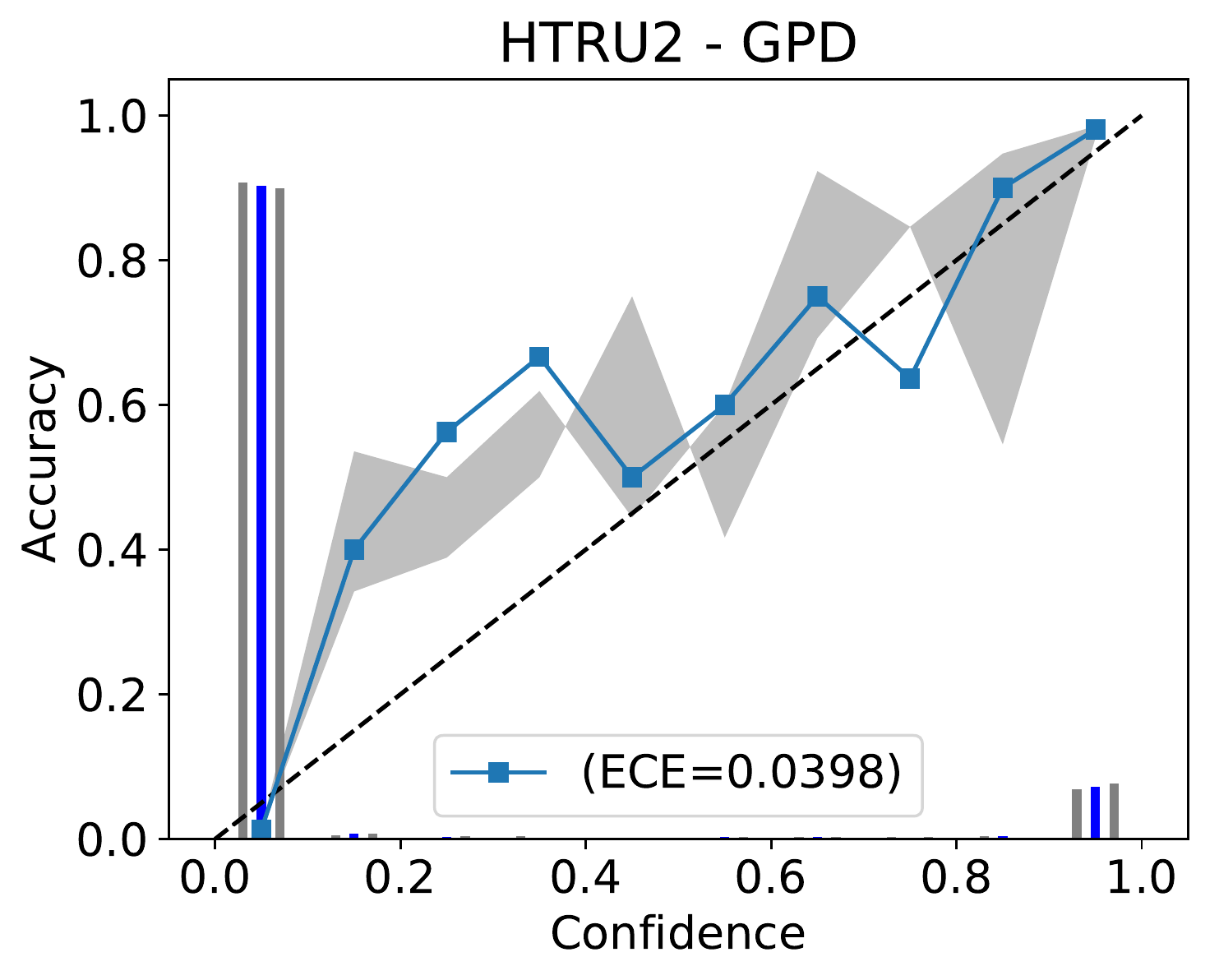}
\includegraphics[width=0.24\linewidth]{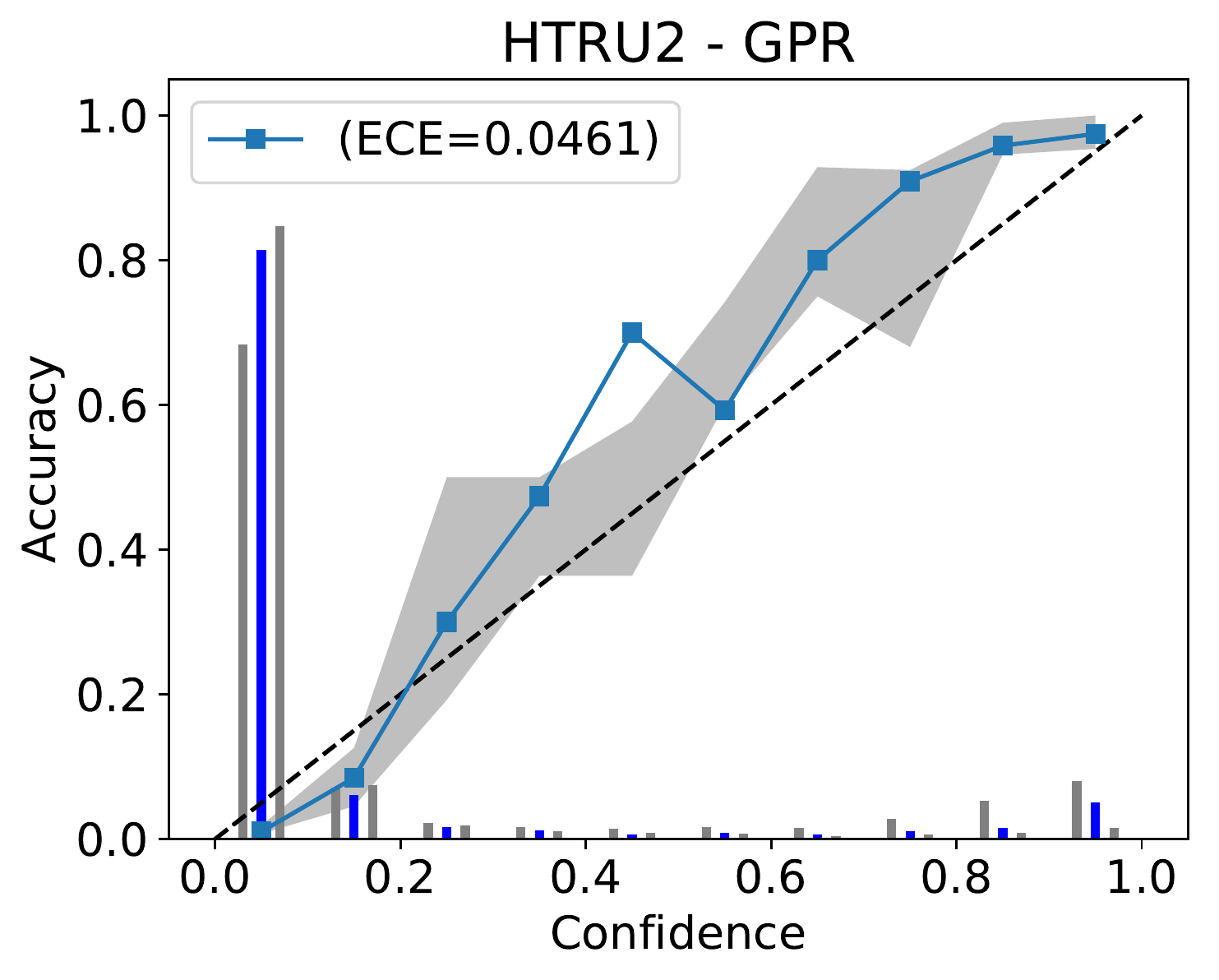}
\includegraphics[width=0.24\linewidth]{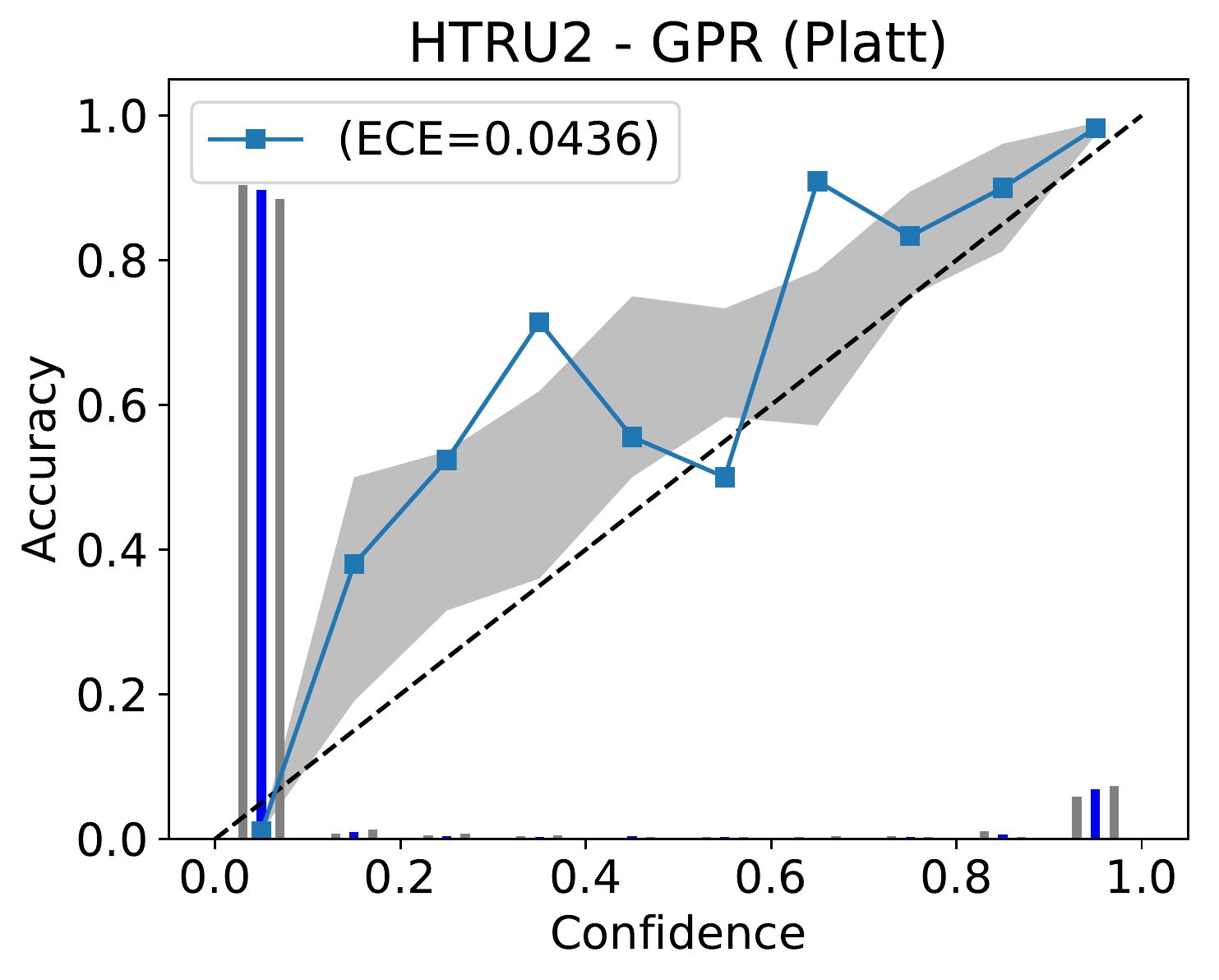}
\includegraphics[width=0.24\linewidth]{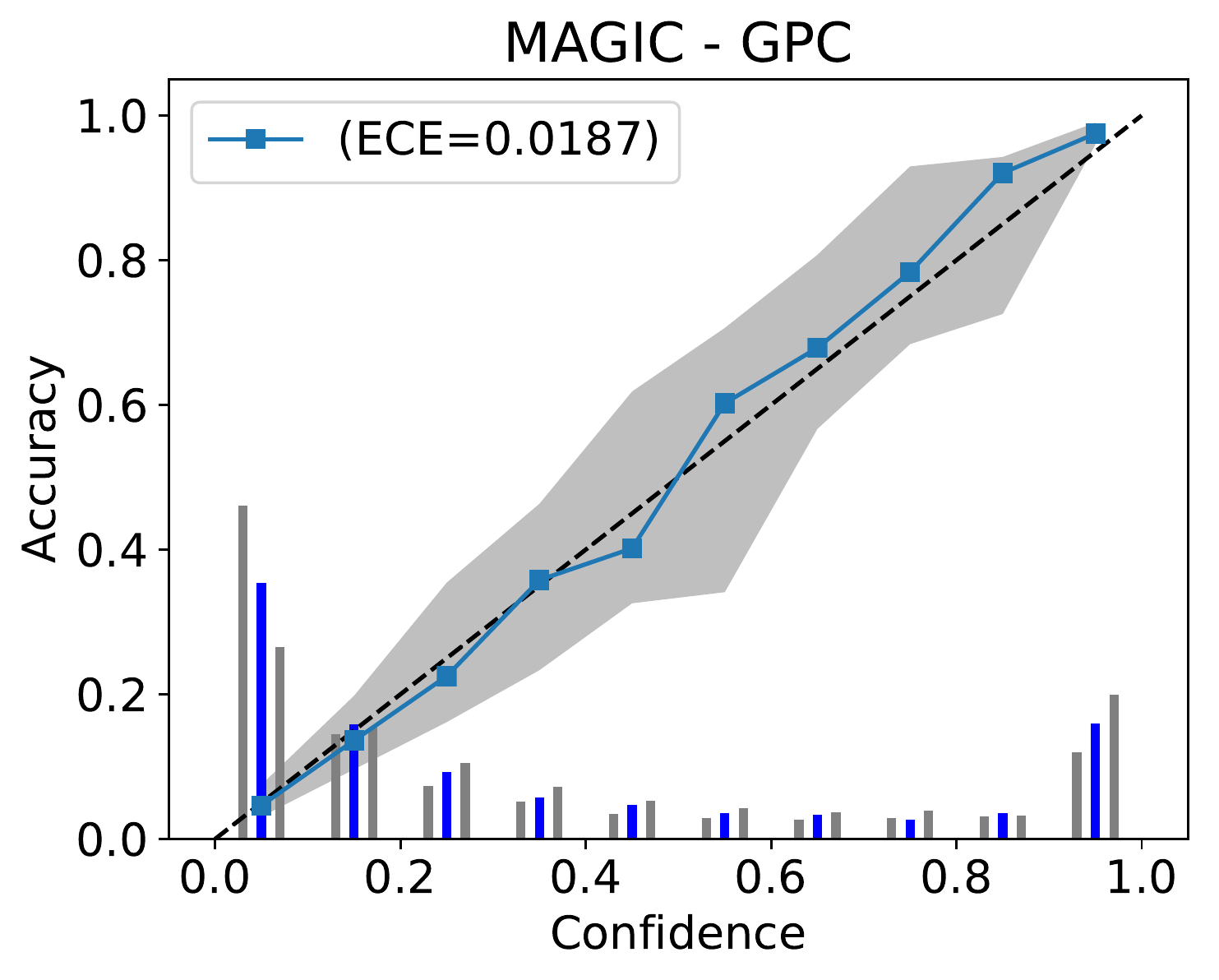}
\includegraphics[width=0.24\linewidth]{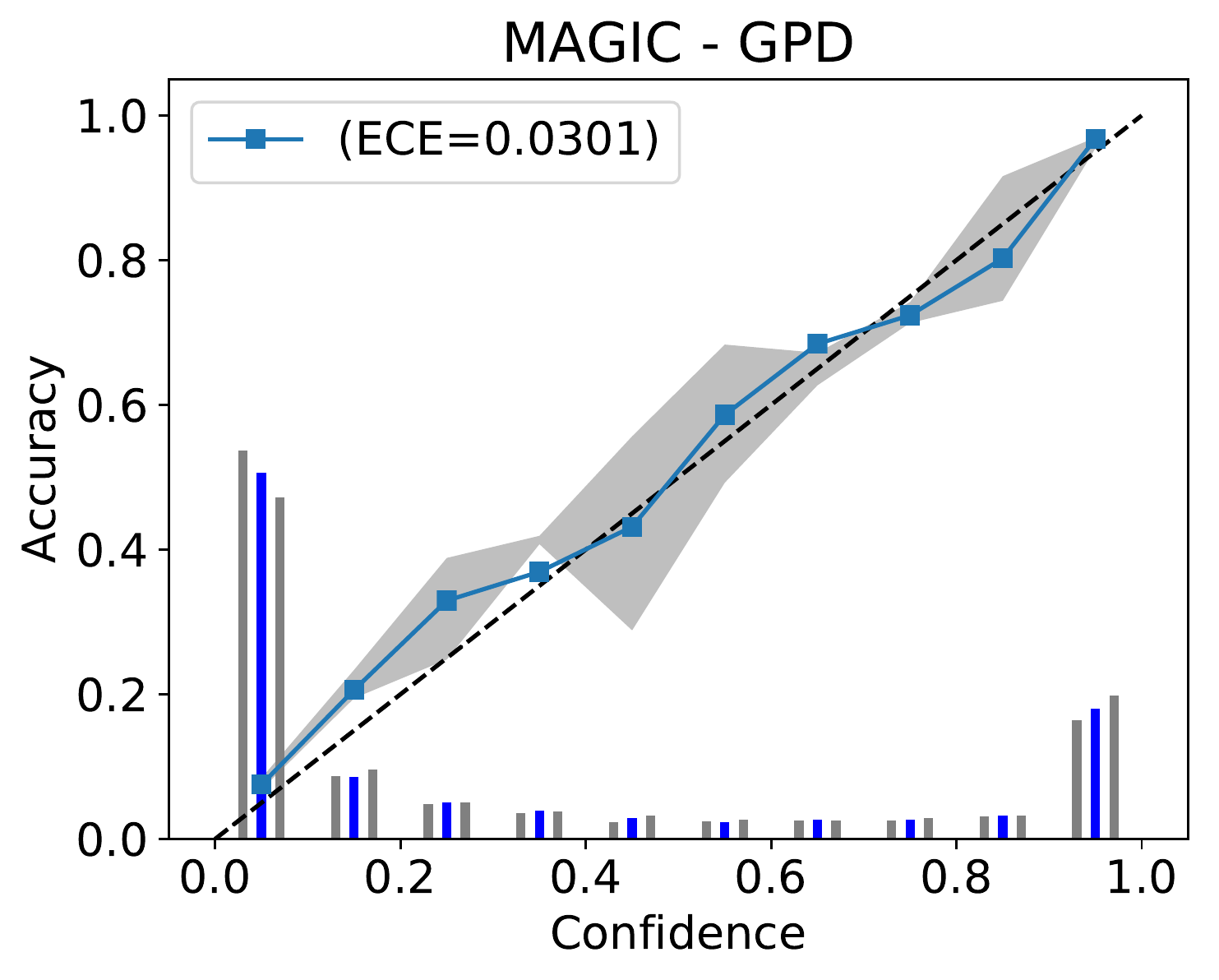}
\includegraphics[width=0.24\linewidth]{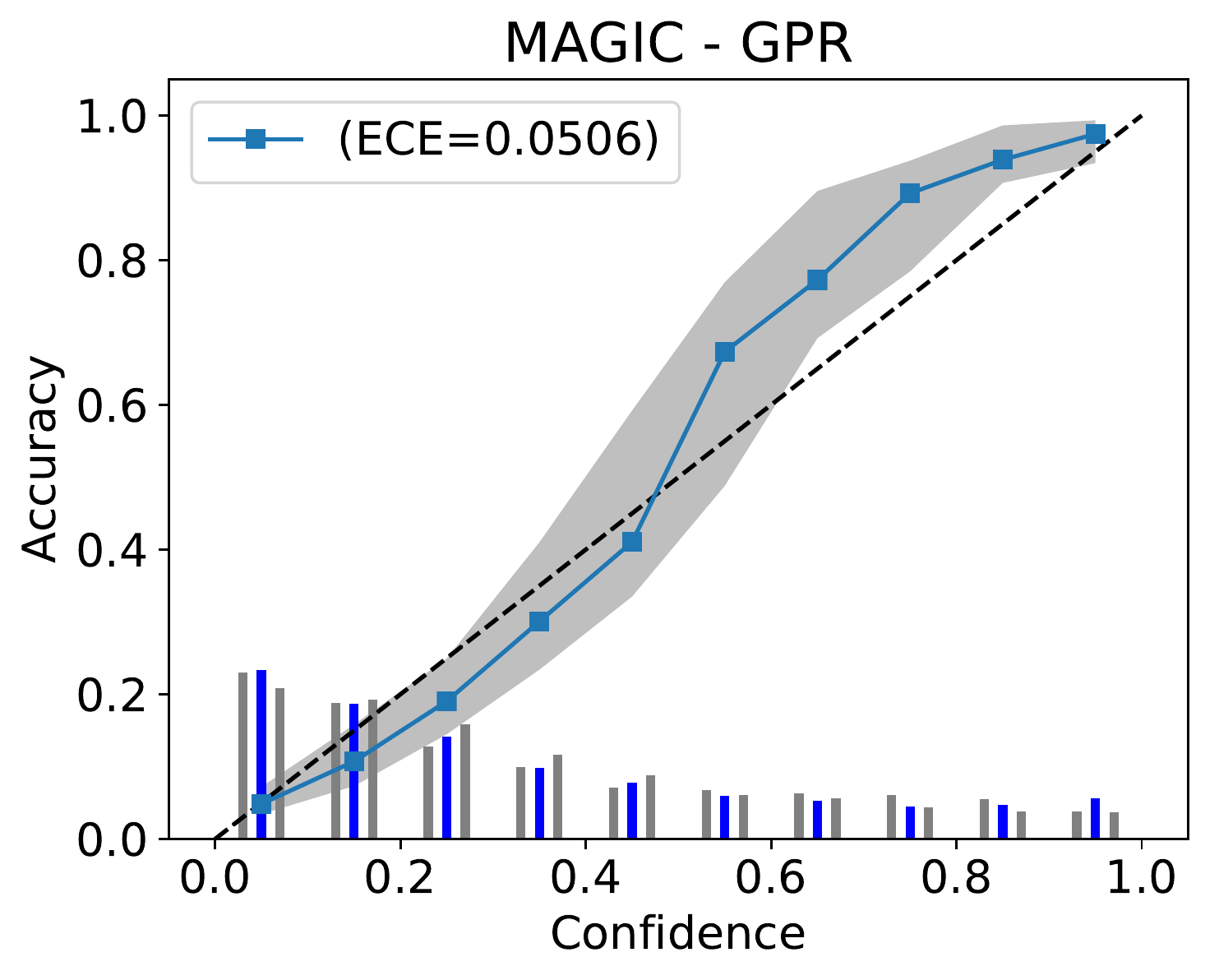}
\includegraphics[width=0.24\linewidth]{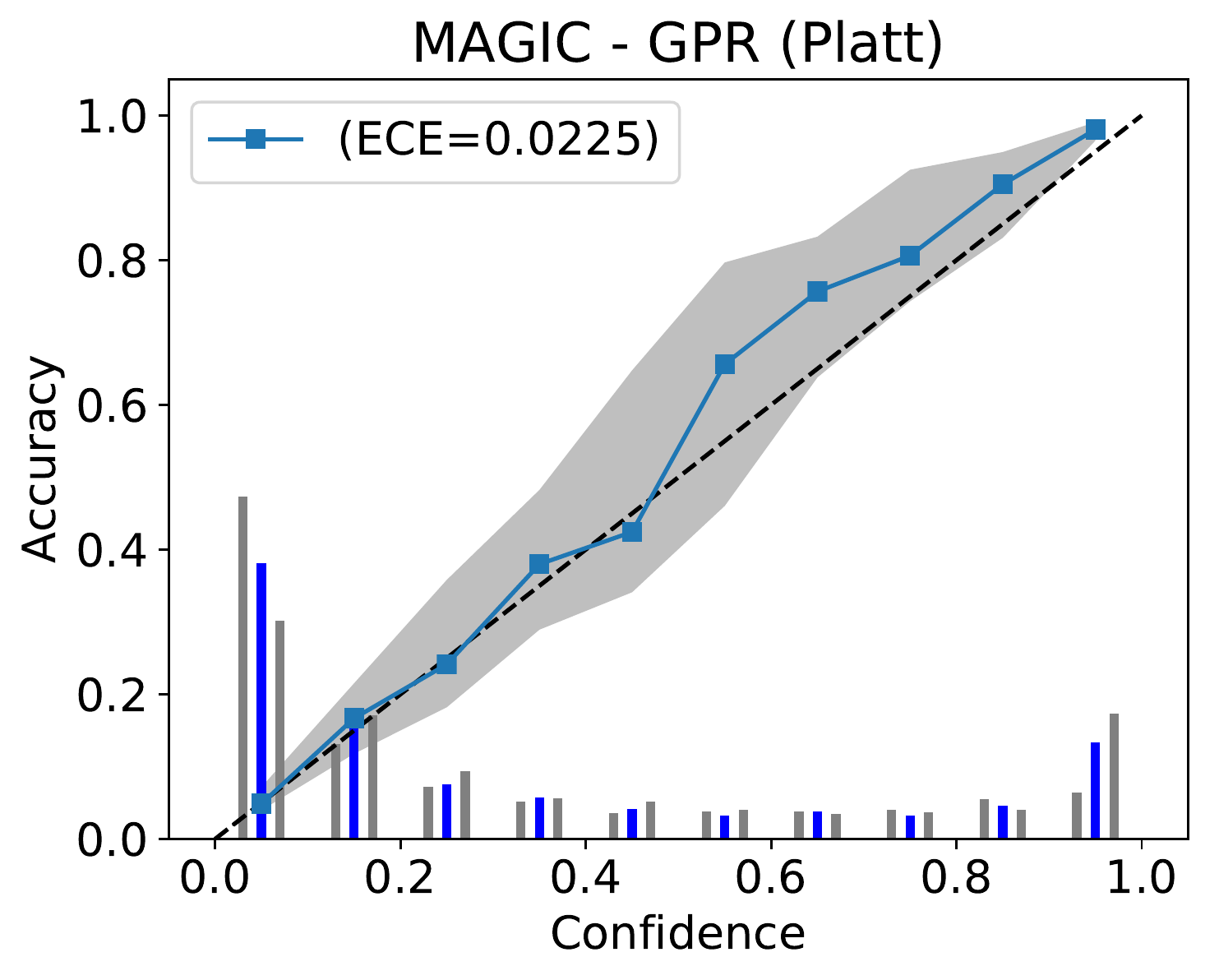}
\includegraphics[width=0.24\linewidth]{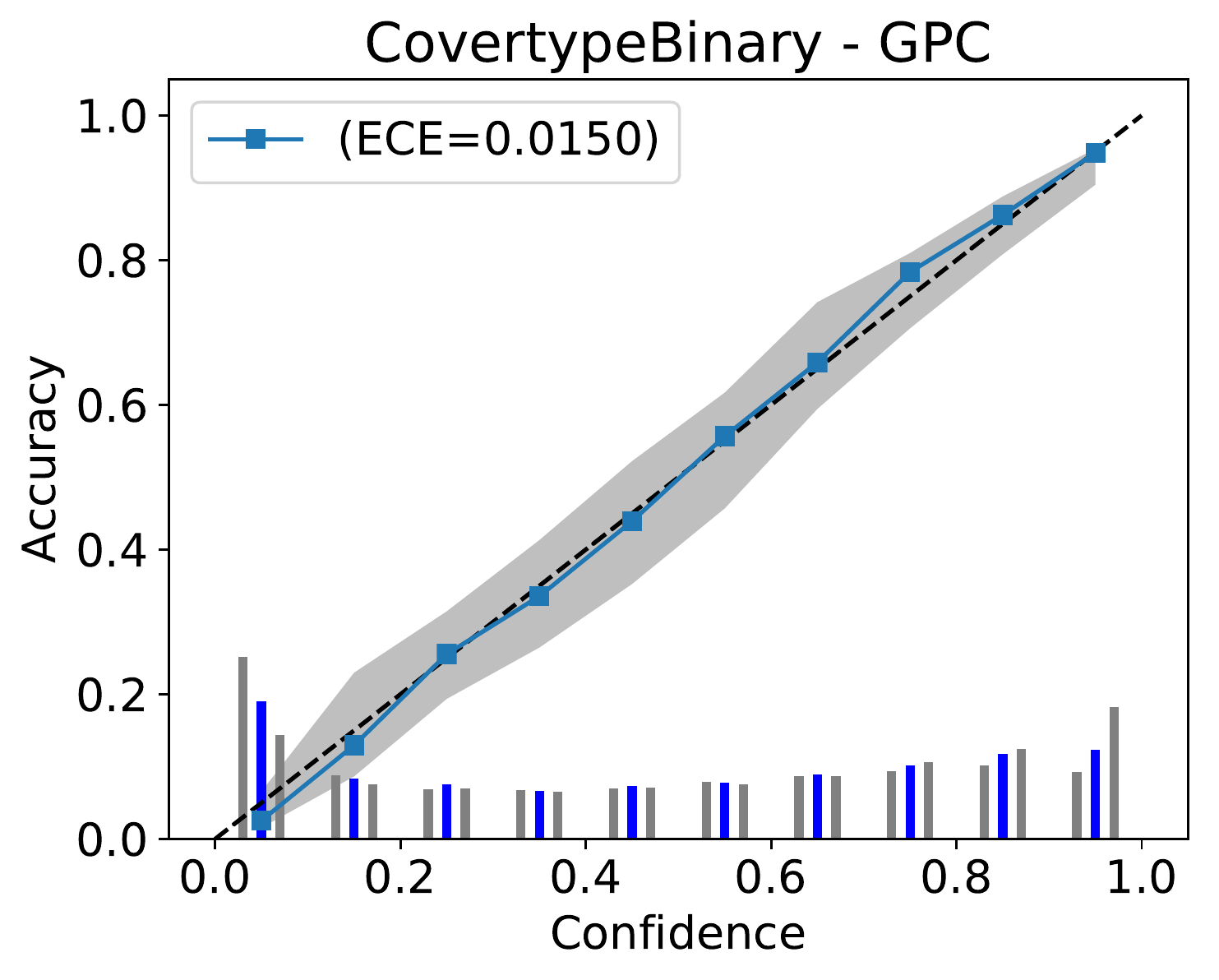}
\includegraphics[width=0.24\linewidth]{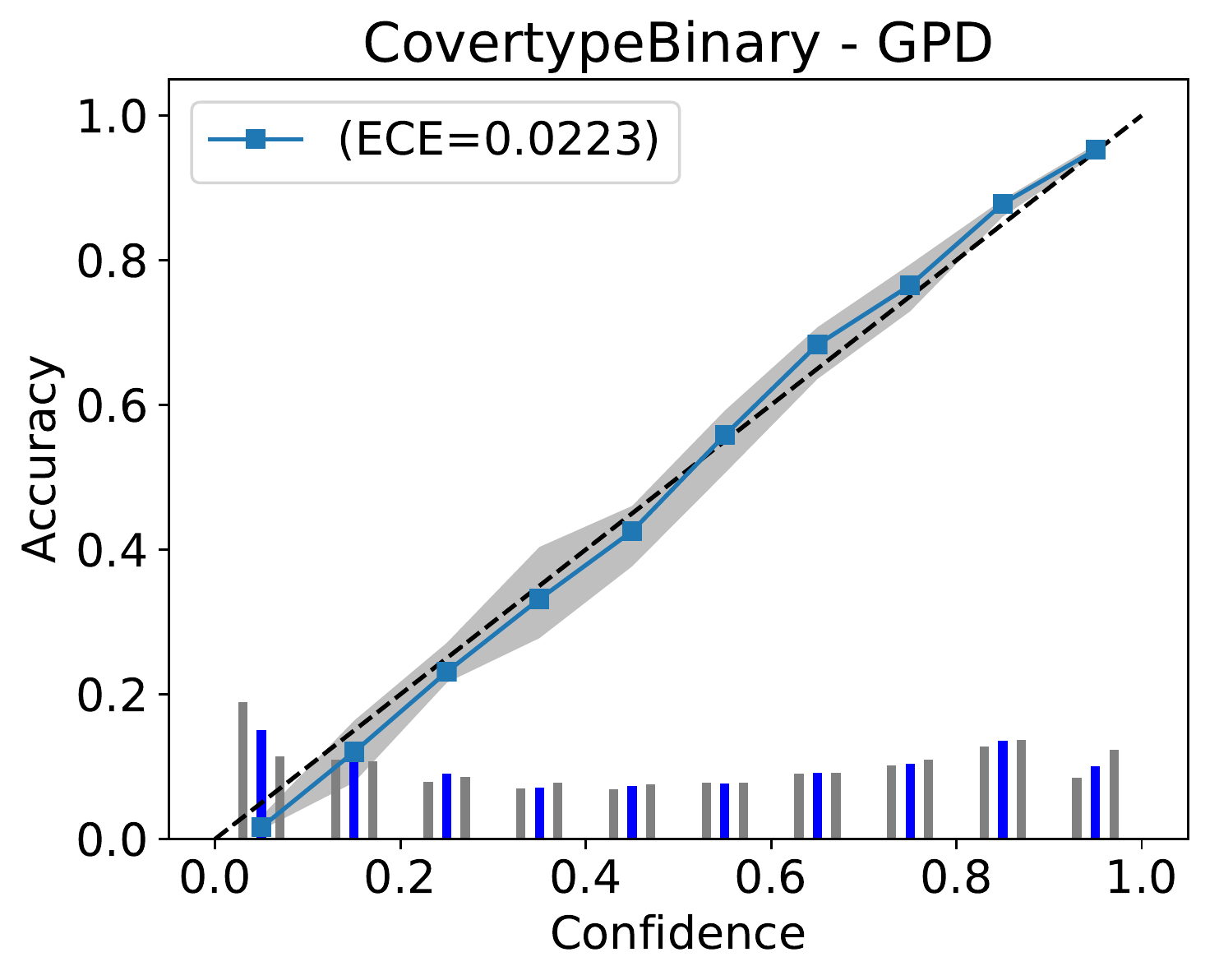}
\includegraphics[width=0.24\linewidth]{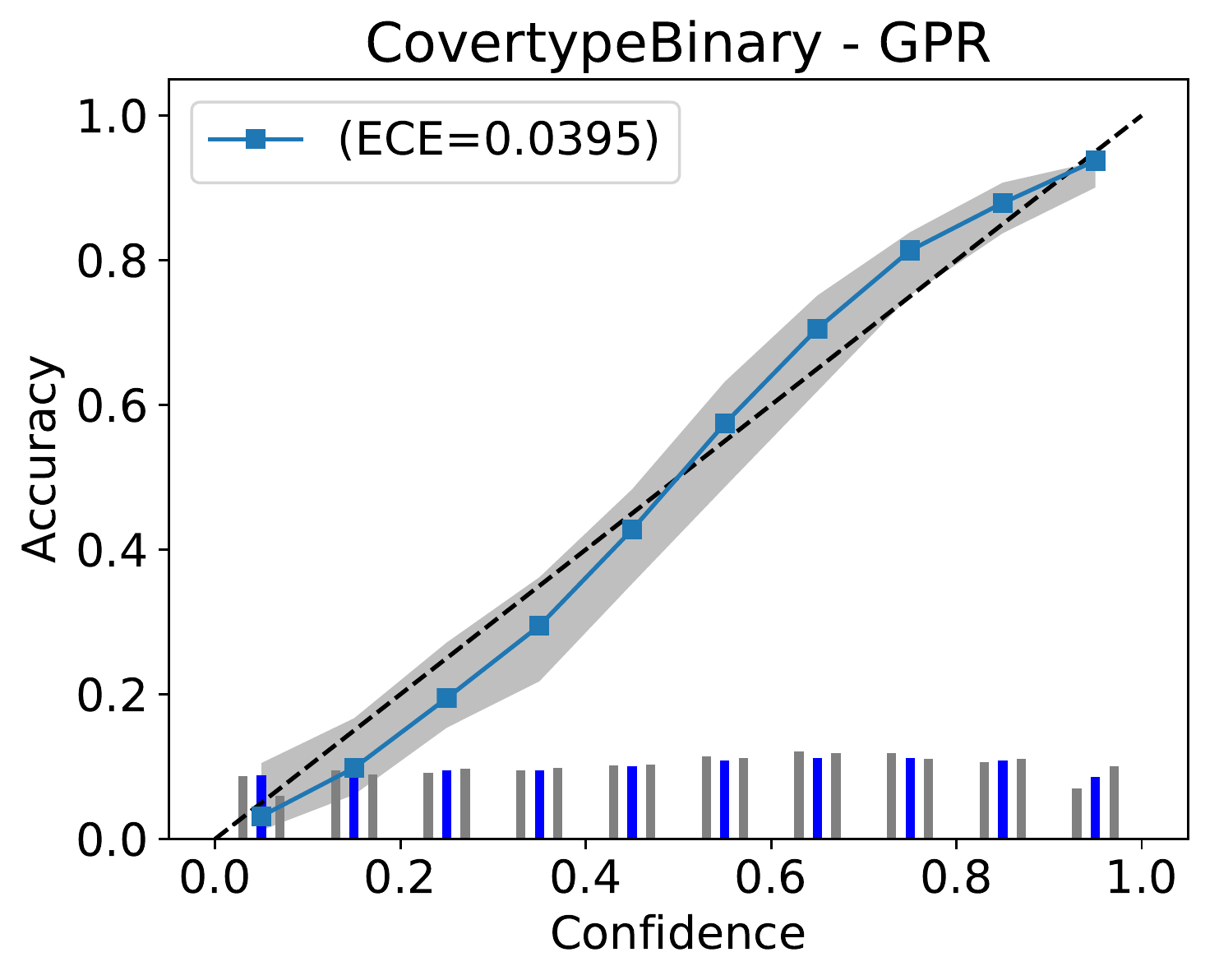}
\includegraphics[width=0.24\linewidth]{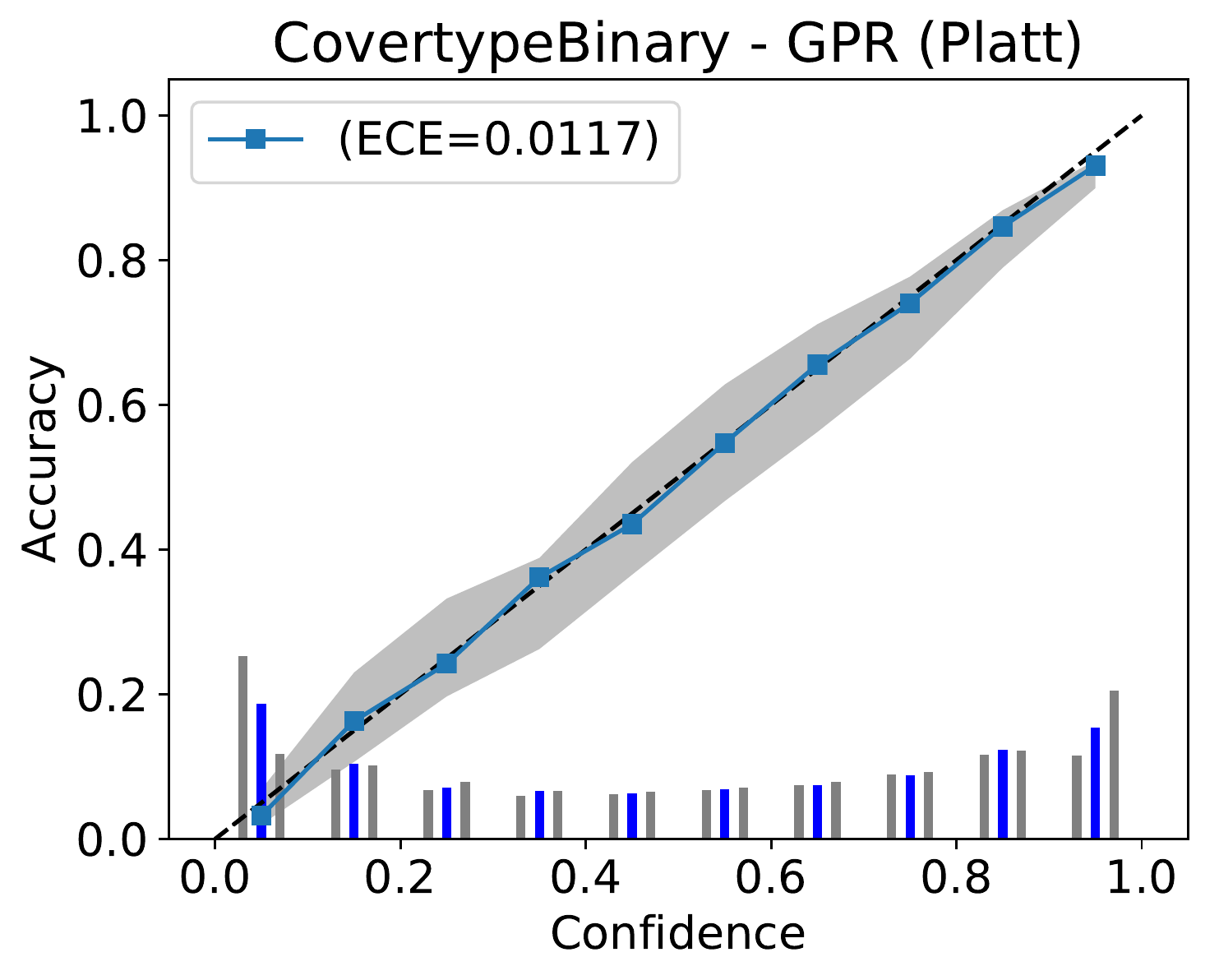}
\caption{Calibration results for four different \gp-based classification approaches on four binary classification datasets. The bounds correspond to the classifiers given by the 95\% confidence interval of the posterior \gp.}
\label{fig:calib_bounds}
\end{figure}

In Figure \ref{fig:calib_bounds} we summarize the reliability diagrams for a number of binary classification datasets.
Each row of diagrams corresponds to a particular dataset; and each column to one of the \gp-based classification approaches that we have discussed in this work.
In particular we consider the variational \gp classification algorithm of \cite{Hensman15} (\gpc), our Dirichlet-based classification scheme (\gpd), and \gp regression on the labels without and with a Platt-scaling post-hot calibration step (\gpr and \gprc).
Note that each one of these approaches produces a distribution of classifiers.
Thus, in the diagrams of Figure \ref{fig:calib_bounds} we show the reliability curve of the mean classifier (depicted as solid lines-points), along with the classifiers described by the upper and lower 95\% quantiles of the predictive distribution (grey area).
If a classifier is well-calibrated, then its reliability curve should be close to the identity curve (dashed line); the latter should also lie within the limits of the grey area.

For the results of Figure \ref{fig:calib_bounds}, we have considered $M=10$ subsets for different levels of confidence.
We note that deviations from the identity curve should not deemed important, if these are not backed by a sufficient number of samples.
For some datasets there are certain levels of confidence (middle section of \htru for example) that contain very few data.
In order to reflect this behavior, we also plot the histograms showing the proportion of the test set that corresponds to each confidence level.

A careful inspection of Figure \ref{fig:calib_bounds} suggests that both \gpc and \gpd produce well-calibrated models.
The \gpr approach on the other hand tends to produce a sigmoid-shaped reliability curve, which suggests that there is underestimation of the class probabilities.
This behavior is cured however by performing calibration via Platt-scaling, as we see for the \gprc method.

\begin{figure}
\centering
\includegraphics[width=0.32\linewidth]{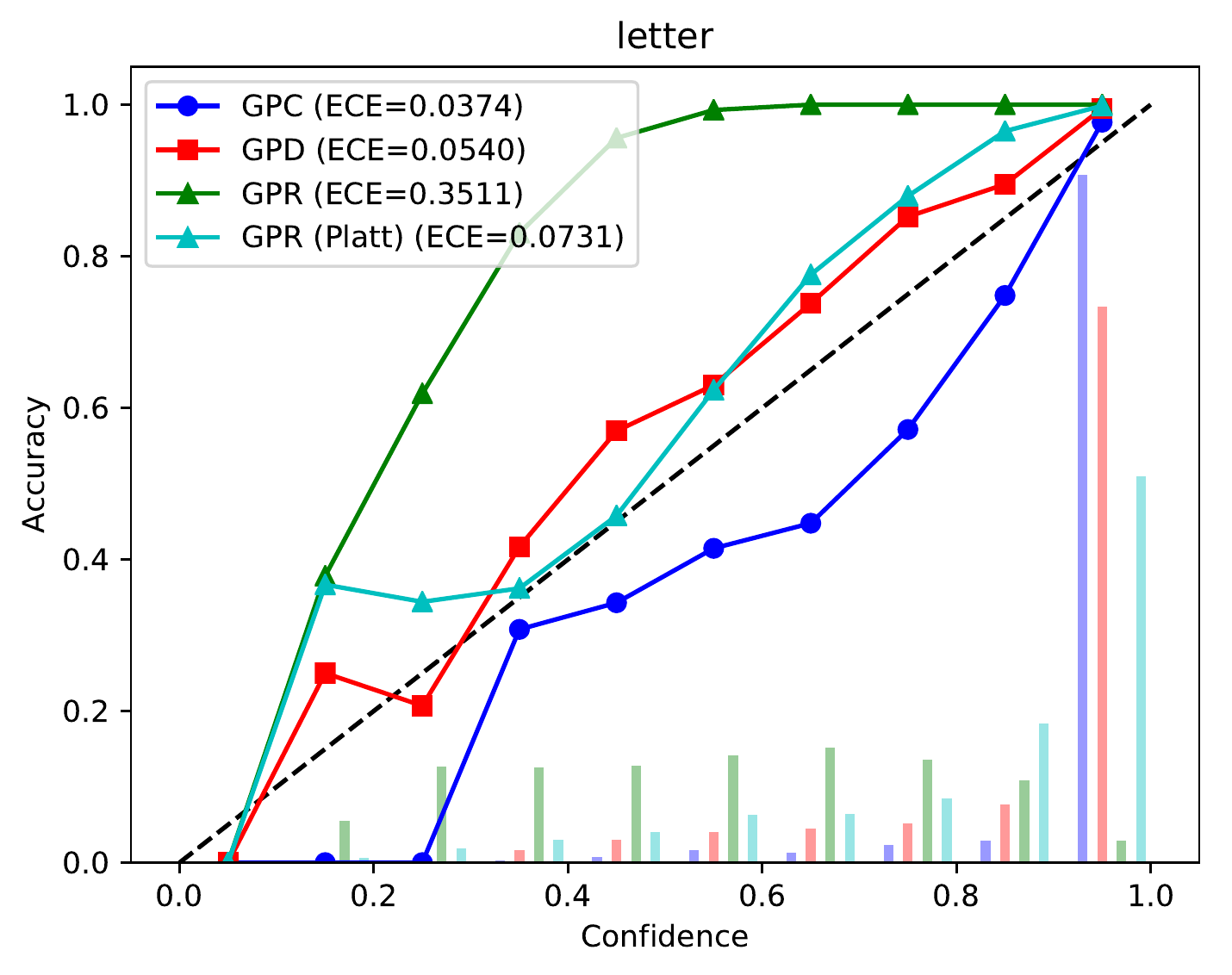}
\includegraphics[width=0.32\linewidth]{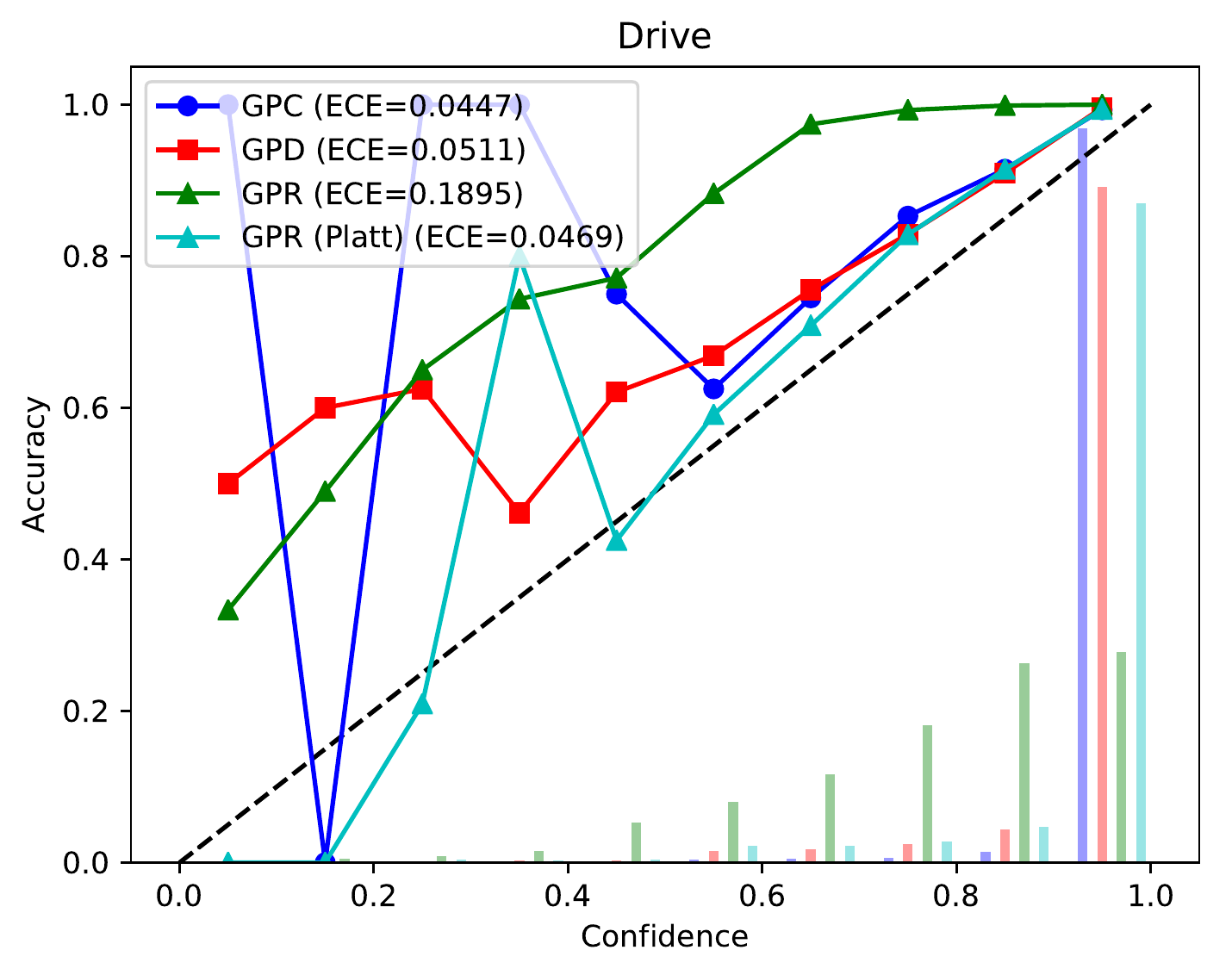}
\includegraphics[width=0.32\linewidth]{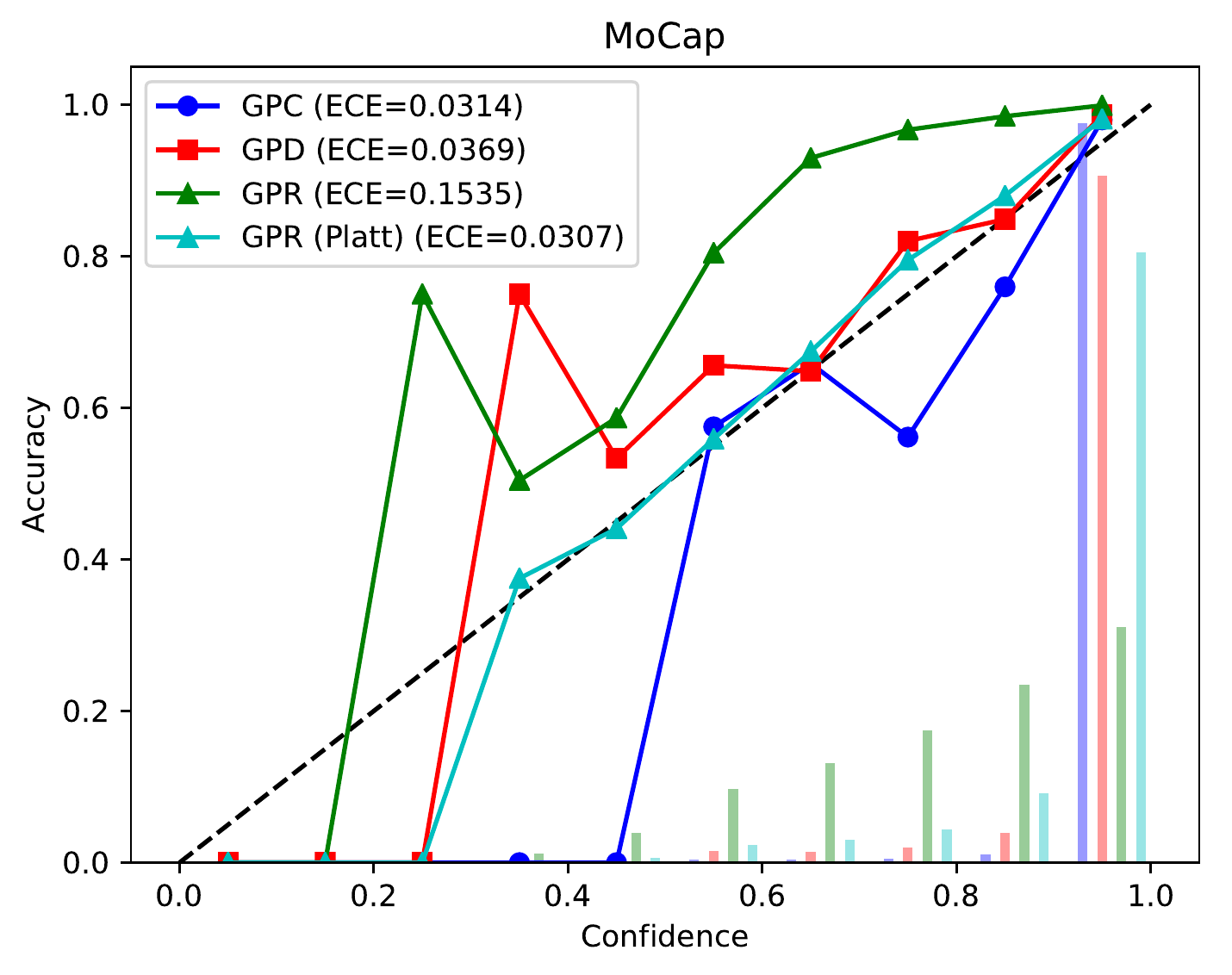}
\caption{Calibration results for three multiclass classification datasets.}
\label{fig:calib_multiclass}
\end{figure}

These conclusions are further supported by Figure \ref{fig:calib_multiclass}, which summarizes the reliability plots for a number of multi-class datasets.
In the multi-class case, it is not obvious how to concisely summarise the effect of the predictive distribution, so we resort to simple reliability plots of the average classifier for each method.

As a final remark, we note that the predictive distribution of the \gpd approach is different from that obtained by \gpc.
In fact, judging form the upper and lower quantile classifiers as presented in Figure \ref{fig:calib_bounds}, it appears that \gpd results in a narrower predictive distribution, which is nevertheless well-calibrated.
An extended characterization of the predictive distribution for \gpd is subject of future work.

\end{document}